%% file: main.tex
\documentclass[letterpaper, 10pt, journal, twoside]{IEEEtran}

\IEEEoverridecommandlockouts 

\makeatletter

\let\proof\@undefined
\let\endproof\@undefined
\makeatother

\usepackage{hyperref}
\hypersetup{
  colorlinks,
  citecolor=black,
  filecolor=black,
  linkcolor=black,
  urlcolor=black,
  pdfauthor={},
  pdfsubject={},
  pdftitle={}
}

\usepackage{graphicx}
\usepackage{setspace}
\usepackage{epstopdf}
\usepackage{float}
\usepackage{multirow,tabularx,makecell}

\usepackage{cellspace}
\newcolumntype{D}{>{\hfill}N{3}{2}<{\hfill}}

\usepackage{wrapfig}
\usepackage[font={small}]{caption}
\usepackage{scalefnt}

\usepackage{subfig}
\usepackage[export]{adjustbox}
\usepackage{environ}
\usepackage[para, bottom]{footmisc}
\usepackage{color}
\usepackage{xcolor}
\usepackage{url}

\usepackage{amsmath}
\usepackage{amssymb,amsmath}
\usepackage{mathtools}
\usepackage{siunitx}
\usepackage[nameinlink, noabbrev, capitalize]{cleveref}
\usepackage{cite}

\usepackage{algorithm}
\usepackage{etoolbox}\AtBeginEnvironment{algorithmic}{\scriptsize}
\usepackage[noend]{algpseudocode}
\usepackage{isotope}
\usepackage{listings}
\makeatletter
\def\lst@makecaption{%
  \def\@captype{table}%
  \@makecaption
}
\makeatother

\makeatletter
\def\BState{\State\hskip-\ALG@thistlm}
\makeatother

\usepackage{tikz}
\usepackage{pgfplots}
\usetikzlibrary{backgrounds,arrows,automata,shapes,positioning,calc,through}
\pgfdeclarelayer{background}
\pgfdeclarelayer{foreground}
\pgfsetlayers{background,main,foreground}

\tikzset{
  state/.style={
    rectangle,
    draw=black, very thick,
    minimum height=1.0em,
    text centered,
  },
  final_state/.style={
    rectangle,
    rounded corners,
    draw=black, very thick,
    minimum height=2em,
    text centered,
  },
  initial_state/.style={
    rectangle,
    double=white,
    double distance=1pt,
    inner sep=2pt,
    draw=black, very thick,
    minimum height=2em,
    text centered,
  },
  point/.style={
    circle,
    inner sep=0pt,
    minimum size=3pt,
    fill=red
  },
  adder/.style={
    circle,
    inner sep=2pt,
    minimum size=0.3in,
    draw=black, very thick,
    text centered
  }
}

\usepackage{soul}

\usepackage{scalerel}
\usetikzlibrary{svg.path}
\definecolor{orcidlogocol}{HTML}{A6CE39}
\tikzset{
  orcidlogo/.pic={
    \fill[orcidlogocol] svg{M256,128c0,70.7-57.3,128-128,128C57.3,256,0,198.7,0,128C0,57.3,57.3,0,128,0C198.7,0,256,57.3,256,128z};
    \fill[white] svg{M86.3,186.2H70.9V79.1h15.4v48.4V186.2z}
    svg{M108.9,79.1h41.6c39.6,0,57,28.3,57,53.6c0,27.5-21.5,53.6-56.8,53.6h-41.8V79.1z M124.3,172.4h24.5c34.9,0,42.9-26.5,42.9-39.7c0-21.5-13.7-39.7-43.7-39.7h-23.7V172.4z}
    svg{M88.7,56.8c0,5.5-4.5,10.1-10.1,10.1c-5.6,0-10.1-4.6-10.1-10.1c0-5.6,4.5-10.1,10.1-10.1C84.2,46.7,88.7,51.3,88.7,56.8z};
  }
}
\newcommand\orcidicon[1]{\href{https://orcid.org/#1}{\mbox{\scalerel*{
        \begin{tikzpicture}[yscale=-1,transform shape]
          \pic{orcidlogo};
        \end{tikzpicture}
}{|}}}}

\title{Autonomous Reflectance Transformation Imaging \\ by a Team of Unmanned Aerial Vehicles}

\author{V\'{i}t Kr\'{a}tk\'{y}$^{\orcidicon{0000-0002-1914-742X}}$, Pavel Petr\'{a}\v{c}ek$^{\orcidicon{0000-0002-0887-9430}}$, Vojt\v{e}ch Spurn\'{y}$^{\orcidicon{0000-0002-9019-1634}}$, and Martin Saska$^{\orcidicon{0000-0001-7106-3816}}$
  \thanks{Manuscript received: September 10, 2019; Revised December 19, 2019; Accepted January 15, 2020.}
  \thanks{This paper was recommended for publication by Editor Jonathan Roberts upon evaluation of the Associate Editor and Reviewers' comments.
  This work was supported by project no. DG18P02OVV069 in program NAKI II, by CTU grant no. SGS17/187/OHK3/3T/13, and by the Grant Agency of the Czech Republic under grant no. 17-16900Y.}
  \thanks{Authors are with the Faculty of Electrical Engineering, Czech Technical University in Prague, Technick\'{a} 2, Prague 6, {\tt\footnotesize\{\href{mailto:kratkvit@fel.cvut.cz}{kratkvit}|\href{mailto:petrapa6@fel.cvut.cz}{petrapa6}|\href{mailto:spurnvoj@fel.cvut.cz}{spurnvoj}|\href{mailto:saskam1@fel.cvut.cz}{saskam1}\}@fel.cvut.cz}. Digital Object Identifier (DOI): see top of this page.}
}

\begin{document}

\newcommand{\PREPRINTYEAR}{2020}
\newcommand{\PREPRINTPUBLISHER}{IEEE}

\onecolumn
\pagenumbering{gobble}
{
  \topskip0pt
  \vspace*{\fill}
  \centering
  \LARGE{%
    \copyright{} \PREPRINTYEAR~\PREPRINTPUBLISHER\\\vspace{1cm}
	Personal use of this material is permitted.
	Permission from \PREPRINTPUBLISHER~must be obtained for all other uses, in any current or future media, including reprinting or republishing this material for advertising or promotional purposes, creating new collective works, for resale or redistribution to servers or lists, or reuse of any copyrighted component of this work in other works.}
	\vspace*{\fill}
}

\twocolumn 
\pagenumbering{arabic}

\markboth{\copyright{} \PREPRINTPUBLISHER, \PREPRINTYEAR. Accepted to IEEE RA-L. DOI: \href{https://doi.org/10.1109/LRA.2020.2970646}{10.1109/LRA.2020.2970646}}{\copyright{} \PREPRINTPUBLISHER, \PREPRINTYEAR. Accepted to IEEE RA-L. DOI: \href{https://doi.org/10.1109/LRA.2020.2970646}{10.1109/LRA.2020.2970646}}

\maketitle


\begin{abstract}
  A Reflectance Transformation Imaging technique (RTI) realized by multi-rotor Unmanned Aerial Vehicles (UAVs) with a focus on deployment in difficult to access buildings is presented in this paper. RTI is a computational photographic method that captures a surface shape and color of a subject and enables its interactive re-lighting from any direction in a software viewer, revealing details that are not visible with the naked eye. The input of RTI is a set of images captured by a static camera, each one under illumination from a different known direction. We present an innovative approach applying two multi-rotor UAVs to perform this scanning procedure in locations that are hardly accessible or even inaccessible for people. The proposed system is designed for its safe deployment within real-world scenarios in historical buildings with priceless historical value.
\end{abstract}


\input{chapters/introduction}
\input{chapters/problem_definition.tex}

\input{chapters/preliminaries}

\input{chapters/rti_scanning.tex}
\input{chapters/experimental_results.tex}
\input{chapters/conclusion.tex}






\bibliographystyle{IEEEtran}
\bibliography{main}


\end{document}

%% file: chapters/introduction.tex
\section{INTRODUCTION}
\IEEEPARstart{R}{eflectance} Transformation Imaging (RTI) is an image-based rendering method widely used by experts in the field of archaeology, restoration and historical science~\cite{Mytum2018, underwater_rti, miles_pitts_pagi_earl_2014, Mytum2017ReflectanceTI, rti_conservation, Saunders:2017:RTI:3136628.3136726, 7467687}.
Based on the set of images with varying known lighting, a representation of an image is produced by RTI, that enables to view a captured object lit from an arbitrary direction and therefore to easily inspect the three-dimensional character of the object without the need to capture thousands of photographs with lighting from all possible directions.

\begin{figure}[h]
  \vspace{-0.26cm}
  \centering
  \subfloat[RTI dome \tiny{}]{
    \includegraphics[width=0.51\columnwidth]{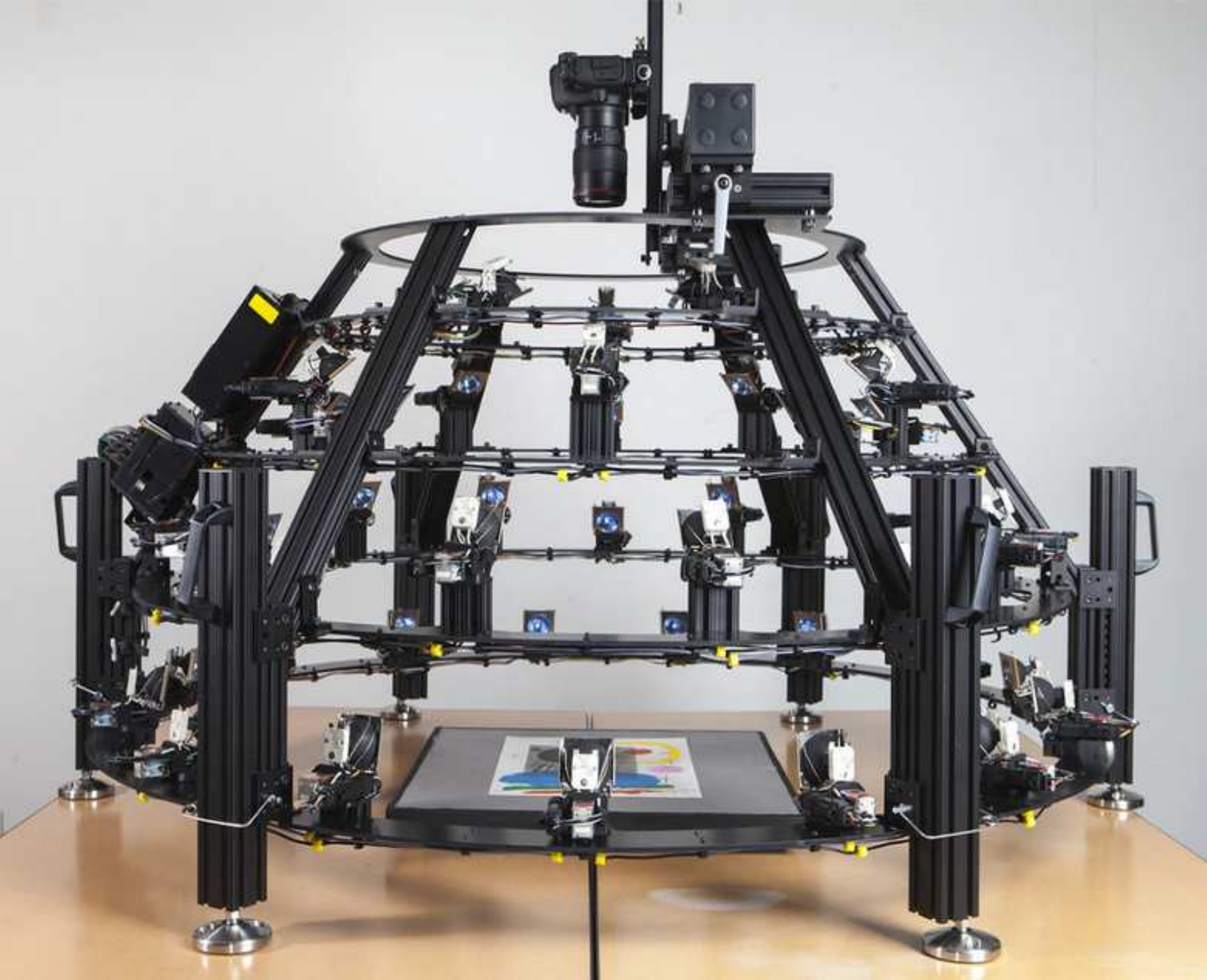}
    \label{fig:rti_dome}
  }
  \subfloat[highlight RTI]{
    \includegraphics[width=0.41\columnwidth]{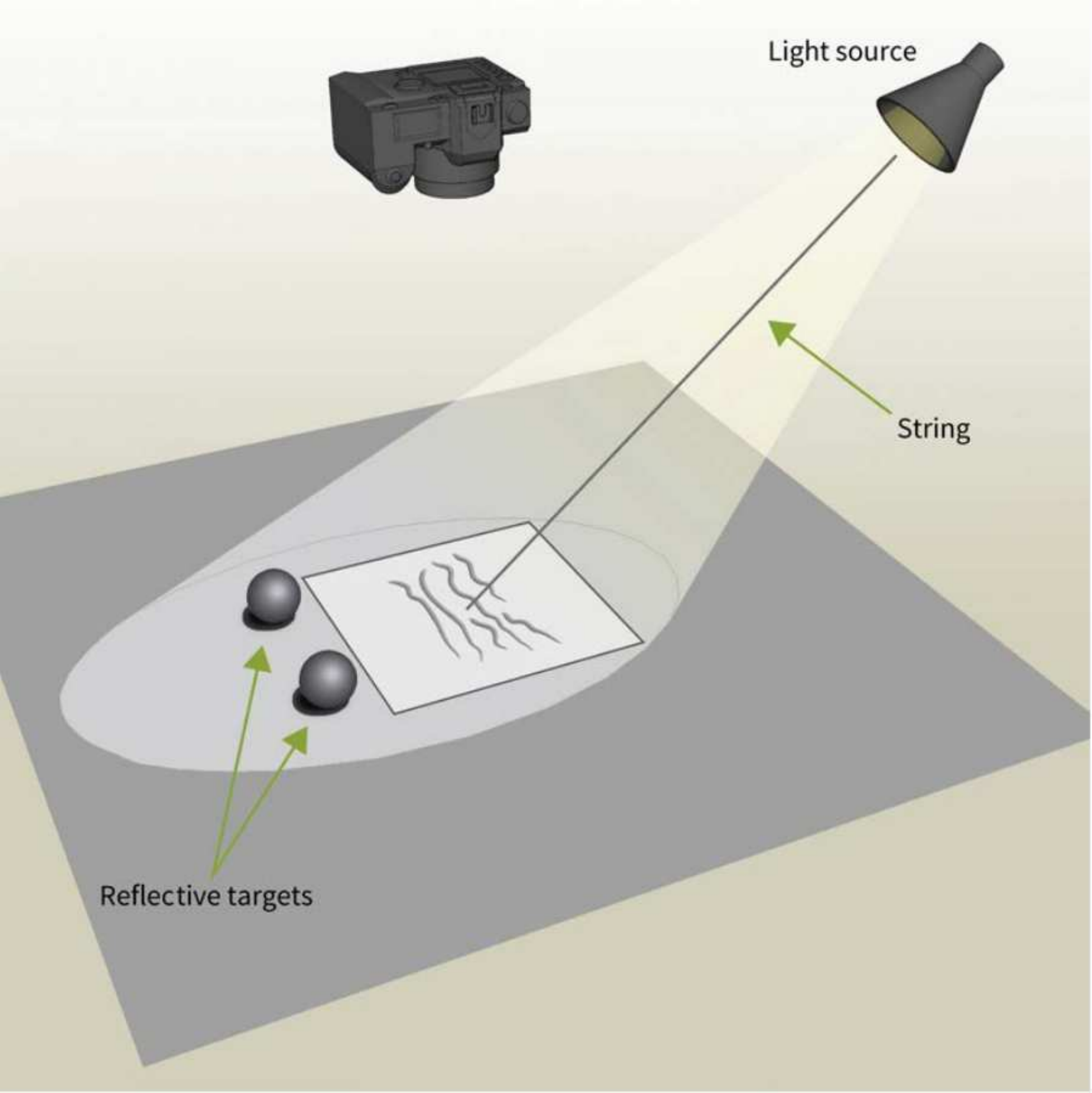}
    \label{fig:hrti}
  }
  \vspace{-0.1cm}
  \caption{Illustration of traditional approaches to the realization of RTI method. \tiny{Image sources: https://www.idigbio.org/wiki/images/7/70/Graham\_0429.pdf and https:// historicengland.org.uk/images-books/publications/multi-light-imaging-heritage-applications/heag069-multi-light-imaging/}}
  \label{fig:rti_approaches}
\end{figure}

The most traditional approaches for gathering the desired set of images is an RTI dome (see~\autoref{fig:rti_dome}) and the Highlight RTI method~\cite{rti_highlight}. The RTI dome includes tens of light-emitting diodes (LEDs) placed on the inner surface of a hemisphere and a camera placed on its top. During the image capturing phase, the RTI dome is placed above the scanned object, and the LEDs are sequentially lit up while the camera is capturing images. Each image is then labelled with the corresponding lighting vector computed from the known position of particular LEDs.

Using the Highlight RTI method (H-RTI), a source of light is manually placed at unknown positions in a constant distance from the scanned object, while a camera mounted on the tripod is capturing images. Respective lighting vectors are then computed from the reflection detected on the high reflective object (metal ball) placed next to the scanned artifact. An example of a setup for the H-RTI method is illustrated in~\autoref{fig:hrti}.

The drawback of both methods is that the scanned object has to be directly accessible to humans, which is difficult to achieve in large historical and sacred buildings. Thus it significantly limits the usage of this very powerful technique. We propose to solve this problem by applying two cooperating multi-rotor UAVs equipped with a camera and light source. The UAV team is able to gather the set of images with corresponding lighting vectors of objects located at places hardly accessible or even inaccessible for people, and much faster than using the H-RTI approach. During the scanning process, the UAV carrying a camera is hovering steadily in the air, while the UAV equipped with a light source is flying around to provide the lighting from different directions.

\begin{figure}[htb]
  \centering
  \includegraphics[width=\columnwidth]{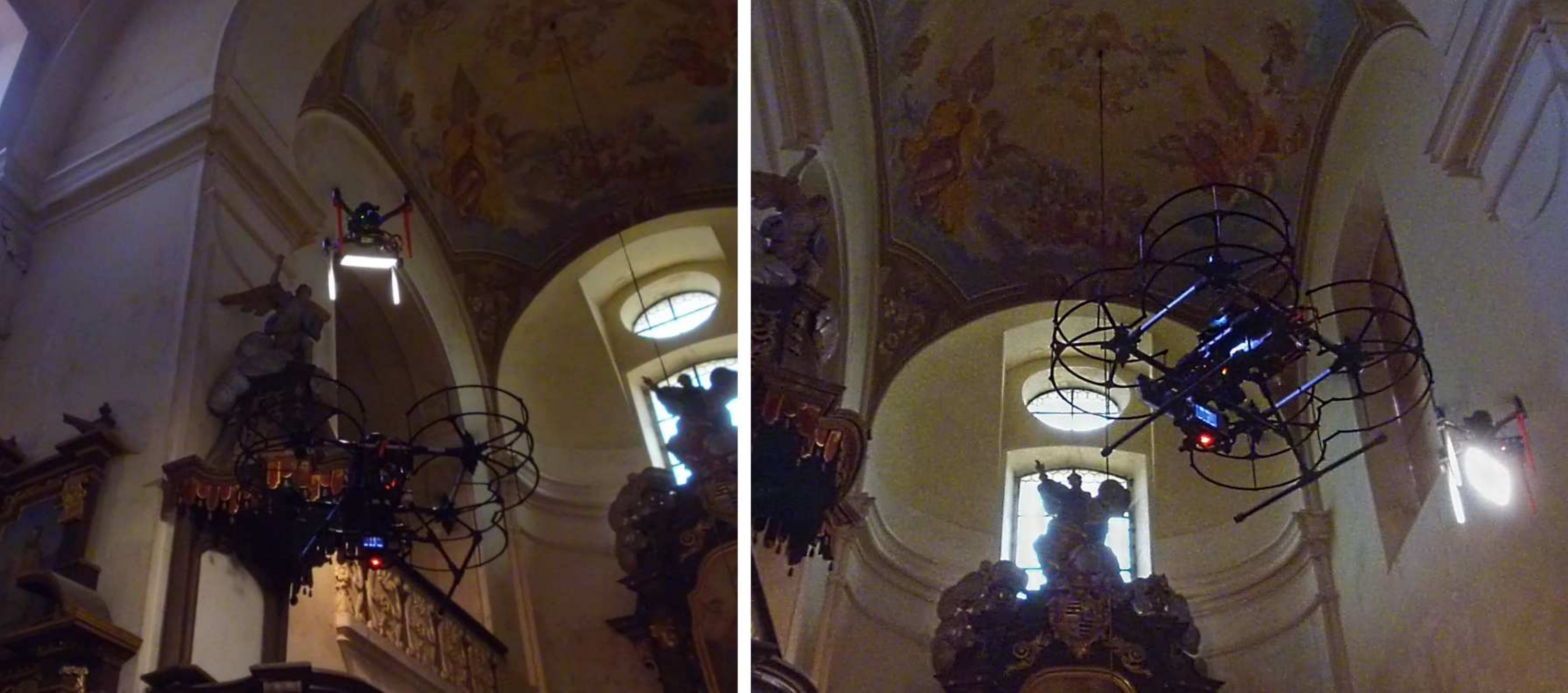}
  \caption{Deployment of the proposed system for autonomous RTI by a team of UAVs in Church of St. Mary Magdalene in \hbox{Chlum\'{i}n}. Multimedia material of the experiment is available at \hbox{\url{http://mrs.felk.cvut.cz/papers/rti2020ral}}.}
  \vspace{-0.7cm}
  \label{fig:real_world_deployment}
\end{figure}

Although the scanning process could be performed manually, regardless of the experience of operators, the manual navigation of UAVs to desired positions is typically less precise than in the case of autonomous control.
Moreover, the desired scanning locations are assumed to be located high above the ground and hence far from the operators, which increases the difficulty and danger of manual control of two UAVs flying close to each other.
Therefore, the proposed method relies on using two fully autonomous and self-localized UAVs.
Although we present the system as autonomous, each UAV is supposed to be monitored by an operator, who is prepared to take over the control in case of an unexpected behavior.
This requirement is given by the aviation authority for flying outdoors (e.g., for scanning statues, mosaics and plasters on the exteriors of churches and castles - see video \hbox{\url{https://youtu.be/lTRqd1gQOAI}}) and by the heritage institute for flying indoors (see~\autoref{fig:real_world_deployment} and videos from Saint Nicholas Church at Old Town Square in Prague
(\url{https://youtu.be/g1NuPnLCFTg}),
Grotto Gorzanow,
Poland (\url{https://youtu.be/6mRYxciDLCM}), and St. Anne's and St. Jacob's Church in Star\'{a} Voda (\url{https://youtu.be/yNc1WfebIag}), where autonomous UAVs have been applied).

The presented application is specific due to cooperation with experts from the Czech National Heritage Institute (https://www.npu.cz/en), who have introduced requirements and constraints, which are untraditional from the robotic point of view. Therefore, two different approaches to the generation of lighting positions and determination of an ideal sequence of these positions to achieve sufficiently good coverage of lighting for the RTI technique are presented in this paper. One of them is based on the method using Fibonacci lattice~\cite{fibonacci_lattice} for achieving approximately equal distribution of points on the sphere, and it applies the approaches for the solution of the Traveling Salesman Problem (TSP) to find a path connecting these positions.
The second proposed approach is aimed to find a compromise among the optimality, robotic constraints, and requirements of the aviation authorities and the heritage institute that require paths producing predictable and easy to follow movement of UAVs, which is optimal regarding the UAV deployment in historical buildings.

  \vspace{-0.3cm}

\subsection{State-of-the-art and contribution}
Single manually controlled UAVs are being commercially used in numerous scenarios, both outdoor and indoor. Nevertheless, the number of possible applications can be significantly increased by introducing autonomous cooperative teams of UAVs, which is the aim of this paper. One of such applications is the documentation of interiors of historical buildings with distributed lighting, which is motivated by the preservation of cultural heritage in the form of digital documentation~\cite{saska17etfa}. It provides the ability to perform later reconstructions of already destroyed historical buildings or art pieces, and also provides the ability to analyze this data and plan the future restoration work without repetitive direct access to particular artifacts.

The documentation of buildings is problematic due to its time complexity and limited accessibility by humans, which naturally leads to the introduction of semi-autonomous or autonomous systems developed for this purpose.
Works related to the scanning of buildings are mostly interested in the planning of the best sensing locations~\cite{roberts:2017, 5979939, 6696481} and only a few of them aim to exploit autonomous vehicles. In~\cite{4399581}, an unmanned ground vehicle (UGV) equipped with a laser scanner capable of autonomous planning of scanning locations and moving through a large scale outdoor environment is introduced. In~\cite{7991427}, authors exploit advantage of UAV systems to operate in larger space by applying them to autonomous inspection of industrial chimneys. Regarding the documentation of particular artifacts in interiors of historical buildings, the only work we have found is~\cite{hallermann}, where technology assisting an operator of a single UAV explicitly developed for this application is described.

Regarding the multi-robot systems works presenting systems deploying UGVs or UAVs for the documentation or mapping of buildings, the robotic groups are focused on reducing the overall mission time or on expanding the scanned area~\cite{8598942, 7152283}, but the direct cooperation of robots is not exploited. The cooperative lighting by a UAV team introduced in~\cite{saska17etfa}, and extended for use of the RTI technique in this paper, is unique in comparison to all the aforementioned works since it employs a team of cooperating UAVs in tasks that cannot be solved by a single UAV only in principle. The proposed method in~\cite{saska17etfa} is exceptional in its approach to actively influence its surrounding environment in order to increase quality and variety of gathered digital material.

In this work, we introduce the first system for autonomous realization of the RTI technique independent on the location of scanned objects, which takes the advantage of our previous works on formation control~\cite{saska17etfa, saska2014jfr, spurny_mmar16, saska_ras15}.

%% file: chapters/problem_definition.tex
\section{PROBLEM DESCRIPTION}

The problem of autonomous realization of Reflectance Transformation Imaging technique consists of 1) determining a set of desired positions of a light source, 2) finding a feasible trajectory so that a UAV can provide illumination from these positions, 3) precise mutual localization of UAVs, and 4) processing the captured images for computation of the desired representation of an image.
The team of UAVs consists of one UAV equipped with a high-resolution camera and one UAV carrying a light source. Both UAVs are assumed to be capable of steady hovering in the air and controlling the orientation of the camera and the light independently of their motion.

We suppose that the UAVs operate in a known environment represented by a map, obtained from a three-dimensional scan of the historical building, and they are equipped with necessary sensors and software for their precise localization and state estimation~\cite{baca_jfr18}. The map of the environment is obtained from a three-dimensional terrestrial laser scanner, providing an incomplete map with missing data in occluded out-of-view locations. Such map is sufficient for localization, however does not provide sufficiently precise and complete models of particular artifacts.
Requirements on the scanning process of an object are given by specification of the RTI technique~\cite{ptm_one} and a position of the object selected for scanning is known prior the mission. Both UAVs are able to accurately follow the trajectory given by the sequence of configurations in an available map of the environment~\cite{baca2018mpc}.

The output of the system is the requested representation of the image computed from the set of images taken with the camera carried by the UAV. Corresponding lighting vectors, which are needed for computation of this representation, are obtained from a known position of the scanned object and positions of the UAV carrying the light.

%% file: chapters/preliminaries.tex
\section{PRELIMINARIES}

\subsection{Reflectance Transformation Imaging}

Reflectance Transformation Imaging (RTI) is an image-based rendering method used for obtaining a representation of an image that enables it to be displayed under arbitrary lighting conditions. 
One type of such representation is the Polynomial Texture Map (PTM), which was proposed by \hbox{T. Malzbender~\cite{ptm_one}}. In contrast to the common representation of an image, where each pixel has assigned three static values for red, green and blue color (RGB), the simplified version of PTM represents the intensity of each color channel $I_{c, x, y},\, c \in \{red,\, green,\, blue\}$ of the pixel at position $(x, y)$ by function
\begin{equation}
    I_{c, x, y} = f(l_u, l_v),
  \end{equation}
  where $l_u$ and $l_v$ are elements of lighting vector and the function $f(\cdot)$ is a second-order bi-quadratic polynomial function with varying coefficients $\alpha_{i, c}$ for particular pixels $(x, y)$. Thus the intensity $I_{c, x, y}$ of each color can be interpreted as
  \begin{equation}\label{eq:rti_biquadratic}
    \begin{split}
    I_{c,x,y} = \alpha_{1, c} l_u^2 + \alpha_{2, c} l_v^2 + \alpha_{3, c} l_u l_v + \alpha_{4, c} l_u + \\
    +\alpha_{5, c} l_v + \alpha_{6, c},\,\, c \in \{red,\, green,\, blue\}.
    \end{split}
  \end{equation}
   The input of the RTI method is a set of images taken from the same viewpoint under varying known lighting conditions, where each image in the set has assigned corresponding lighting vector. With the use of this data, coefficients in equation~\eqref{eq:rti_biquadratic} can be computed for all pixels and their color channels (see~\cite{ptm_one} for details).

\subsection{Localization}

Precise determination of position and orientation of UAVs is a crucial assumption for the good performance of the introduced documentation method. Since we aim at the deploying of the system mostly in indoor environments, we rely on the approach presented in~\cite{icra20petracek}, which is capable of working in environments without a sufficient signal from Global Navigation Satellite Systems (GNSS). The method requires UAVs to be equipped with one $360^\circ$ laser scanner (such as a lightweight RP-Lidar), and two distance sensors (e.g., Garmin LIDAR-Lite v3) oriented downwards and upwards with respect to the frame of UAV. A combination of Iterative Closest Point (ICP) and particle filter algorithm is applied to find the position and orientation of the UAV relative to a three-dimensional point cloud of the environment obtained from a terrestrial laser scanner.

%% file: chapters/rti_scanning.tex
\section{DISTANT AUTONOMOUS RTI METHOD}\label{sec:rti_scanning}

Methods designed for the realization of the RTI scanning technique by a team of UAVs, which are described in the following sections, are highly influenced by the requirements of experts from the field of restoration and historical science, where key factors are safety and deployability independently to an external infrastructure.

\subsection{Generation of the set of lighting positions}
To achieve a good coverage of lighting to a general object during RTI scanning, the lighting vectors need be uniformly distributed over the range defined by the minimum and maximum lighting angles in horizontal ($\lambda_{h, min}$, $\lambda_{h,max}$) and vertical ($\lambda_{v,min}$, $\lambda_{v,max}$) direction. The intensity of lighting presented at the scanned object should be the same for all lighting directions.

Given these two requirements and assumption that the intensity of the light source is constant, we can determine that the desired positions of light sources are distributed on a cap of the sphere with its center located at the position of the scanned object. This task can be defined as the problem of uniform distribution of points on the sphere. Since this problem has an exact solution only for particular cases~\cite{Saff1997}, we apply an approximate approach based on the Fibonacci lattice. Inputs of this process are the number of desired lighting positions to be uniformly distributed over the area defined by angles $\lambda_{d, m}, \, d \in \{h, v\},\, m \in \{min, max\}$, position of an Object of Interest (OoI), and the desired lighting distance. The resulting set of points $\Lambda_c$ computed within this process is constructed as
\begin{equation}
  \Lambda_c = \Lambda \cup P_{i},
\end{equation}
where $\Lambda$ is the set of desired lighting positions and \mbox{$P_{i} \in \mathbb{R}^3$} is the initial position of the UAV carrying the light.

\subsection{Determination of the optimal sequence}

An optimal closed path connecting all the desired RTI lighting positions in the set $\Lambda_c$ with respect to a certain criterion (minimum energy, shortest path, minimum time) needs to be found. This problem can be defined as TSP, which is usually solved by splitting it into two subproblems - finding paths between all possible pairs of positions from the set $\Lambda_c$ and finding the optimal sequence of these paths with respect to a certain criterion. The final path is then given as a connection of paths in the optimal sequence. Using this approach, it is difficult to guarantee feasibility of a composed path with respect to constraints given by the kinematic model of a moving robot. Nevertheless, the RTI method requires a static illumination while capturing an image, and so the UAV carrying the light has to be static for taking each picture in the sequence. Therefore, the UAV should stop at every position from $\Lambda_c$ and the problem of an unfeasible path in connections of curve segments does not need to be considered here.

Considering the expected application of the system, we propose to use the minimum energy as the optimization criterion for the solution of proposed alternative of TSP, which also leads to maximization of possible flight time. Based on our experiments, the energy consumption along the closed trajectory flown at constant velocity is proportional to the length of this trajectory and does not depend on the direction of flight. By combining the observations mentioned above and considering an obstacle-free environment, the problem of finding the optimal sequence of the lighting positions is completely defined as the Euclidean TSP (ETSP). For solution of this problem, we have applied the solver using Lin-Kernighan heuristic~\cite{lkh} (LKH solver), which belongs to the most efficient approximate algorithms for solution of TSP. An example of path produced with the described approach (further referenced as Fib-LKH) is presented in~\autoref{fig:rti_comparison_fib}.

\subsection{Safety pilot predictable approach (SPPA)}\label{sec:alternative_approach}
In this section, we present an alternative approach to the obtaining of scanning plan that aims at the generation of lighting positions close to uniform distribution and finding a short path connecting these positions while complying to requirements on human predictability of the resulting trajectory. Thus this method decreases the time needed for recognition of the faulty behavior by a safety pilot, who monitors the UAV during the scanning process. This approach is motivated by a technique used by restorers during the manual acquisition of images for RTI method in~\cite{DCCS06}.

\begin{figure}[h]
  \vspace{-0.5cm}
  \centering
  \hspace{-0.2cm}
  \subfloat[illustration]{
    \includegraphics[width=0.30\columnwidth]{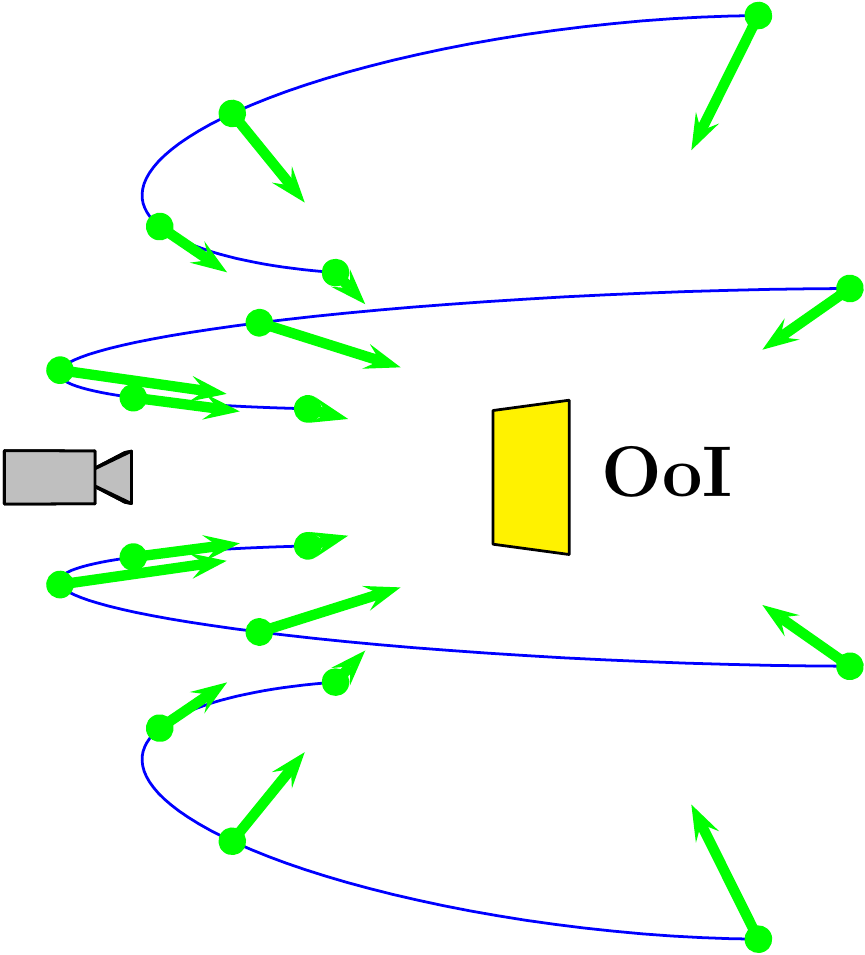}
    \label{fig:rti_horizontal_splines}
  }
  \hspace{-0.2cm}
  \subfloat[experiment in real environment]{
    \includegraphics[width=0.625\columnwidth]{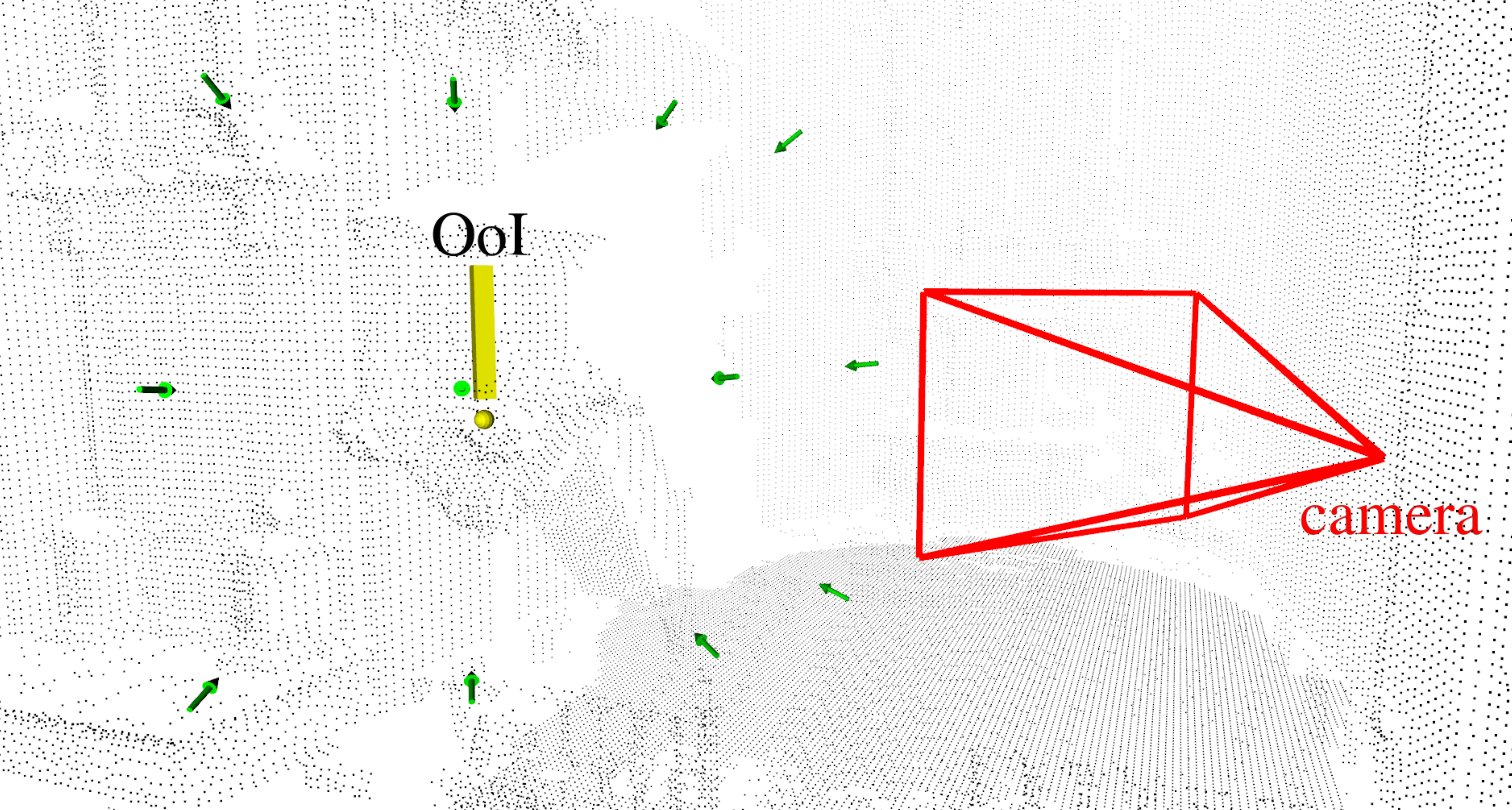}
    \label{fig:rti_rviz}
  }
  \caption[The example of the generated set of RTI goals]{The example of the generated set of RTI goals marked with green dots and arrows. The yellow rectangle identifies the scanned object, and the blue curves indicate the horizontal splines, that represent possible positions of the RTI goals.}
  \vspace{-0.2cm}
\end{figure}

The proposed method for obtaining the set of desired lighting positions uses as inputs the border lighting angles $\lambda_{d, m}, \, d \in \{h, v\},\, m \in \{min, max\}$, the position of scanned object $P_{OoI} \in \mathbb{R}^3$, the orientation of camera defined with yaw and pitch angle $(\psi_{cam}, \zeta_{cam})$, the desired distance between the light and scanned object $d_l$, and the desired number of samples of the lighting angles $v_s$ in vertical direction for which holds $v_s \geq 2$. In the first step of the method, a set of samples of vertical lighting angles $\Lambda_v$ from interval $\langle \lambda_{v, max},\, \lambda_{v, min}\rangle$ is obtained so that they are equally distributed over this interval, $|\Lambda_v| = v_s$, $\min(\Lambda_v) = \lambda_{v, min}$ and $\max(\Lambda_v) = \lambda_{v, max}$. Subsequently one spline on which possible positions of the light lie is constructed for each sample $\lambda_v$ from $\Lambda_v$. These splines are parts of a circle and with given $P_{OoI} = [x_{OoI},y_{OoI},z_{OoI}]^T$ are defined as
\begin{equation}\label{eq:rti_vertical_spline_complete}
  \begin{aligned}
  &\begin{multlined} x_s = x_{OoI} - d_l \cos(\lambda_v + \zeta_{cam}) \cos(\lambda_h + \psi_{cam}),\,\,\\
    \lambda_h \in \langle \lambda_{h, min}, \lambda_{h, max} \rangle
  \end{multlined} \\
  &\begin{multlined} y_s = y_{OoI} - d_l \cos(\lambda_v + \zeta_{cam}) \sin(\lambda_h + \psi_{cam}),\,\,\\
    \lambda_h \in \langle \lambda_{h, min}, \, \lambda_{h, max} \rangle
  \end{multlined}\\
  &z_s = z_{OoI} - d_l \tan(\lambda_v + \zeta_{cam}).
  \end{aligned}
\end{equation}
The splines defined by equation~\eqref{eq:rti_vertical_spline_complete} are graphically illustrated in~\autoref{fig:rti_horizontal_splines}.
The desired distance between the lighting positions on one spline $s_d$ is determined by the equation
\begin{equation}
  s_d = \frac{d_l (\lambda_{v, max} - \lambda_{v, min})}{v_s},
\vspace{0.0cm}
\end{equation}
which corresponds to the shortest distance between two neighboring splines traveled on the surface of the spherical cap. The number of lighting positions on each spline is defined as
\begin{equation}\label{eq:nof_horizontal_samples}
  n_s(\lambda_v) = 1 + \left\lfloor\frac{d_l \cos(\lambda_v) (\lambda_{h, max} - \lambda_{h, min})}{s_d}\right\rceil.
\end{equation}
The set of sample positions $\Lambda_h(\lambda_v)$ on spline corresponding to angle $\lambda_v$ is obtained as
\begin{equation}
  \Lambda_h(\lambda_v) = \left\{\lambda_{h, min} + \frac{\lambda_{h, max} - \lambda_{h, min}}{2}\right\},
\end{equation}
if $n_s(\lambda_v) = 1$ and
\begin{equation}
  \begin{split}
    \Lambda_h(\lambda_v) = \left\{\lambda_{h, min} + k \frac{\lambda_{h, max} - \lambda_{h, min}}{n_s - 1} \right|\\
    k \in \{0, 1, \dots, n_s-1\}\bigg\},
  \end{split}
\end{equation}
if $n_s = (\lambda_v) \geq 2$.
The complete set of the desired lighting positions $\Lambda_c$ generated with the SPPA is defined as
\begin{equation}
  \Lambda_c = \{\Lambda_h(\lambda_v)| \lambda_v \in \Lambda_v\} \cup P_{i}.
\end{equation}
As the first step of the SPPA, the current position of the UAV carrying the light is added at the beginning of the ideal sequence of the RTI positions $S_p$. Then the closest pair of the RTI positions \mbox{$P_{s},\,P_{e} \in \mathbb{R}^3$} on the vertical boundaries needs to be found to select the higher one as the start point and the lower one as the end point among RTI positions. For $P_{s},\,P_{e}$ holds
\looseness=-1
\begin{align}
  P_{s}, P_{e} =
  \arg \min_{P_{i, j}, P_{k, l}} \text{dist}&(P_{i,j}, P_{i}) + \text{dist}(P_{k, l}, P_{i}),\nonumber\\
  s.t.\,\,\phantom{(j, )} i &= k+1, \\
  (j, l) &\in \{(1, 1), (|\lambda_{h, i}|, |\lambda_{h,k}|)\},\nonumber
\end{align}
where $P_{i, j}$ stands for the RTI position in the $i$-th row and $j$-th column, function $\text{dist}(\cdot)$ returns the Euclidean distance between two positions given as arguments, and $\lambda_{h, i}$ stands for the set of RTI positions in the $i$-th row. The position $P_{s}$ is then added to the sequence of positions $S_p$. After that, all positions on the vertical boundary on the way up to the highest row are added to $S_p$. By these three steps, one of the corner positions in the highest row is reached. In the following stages, the procedure depends on the number of rows.

\begin{figure}[h]
  \centering
  \subfloat[]{
    \includegraphics[width=0.28\columnwidth]{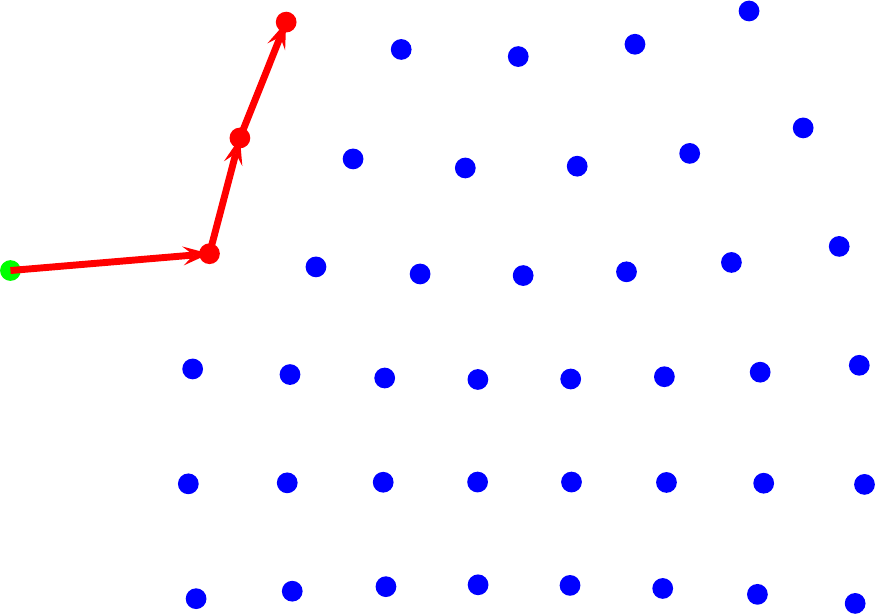}
    \label{fig:rti_even_3}
  }
  \subfloat[]{
    \includegraphics[width=0.28\columnwidth]{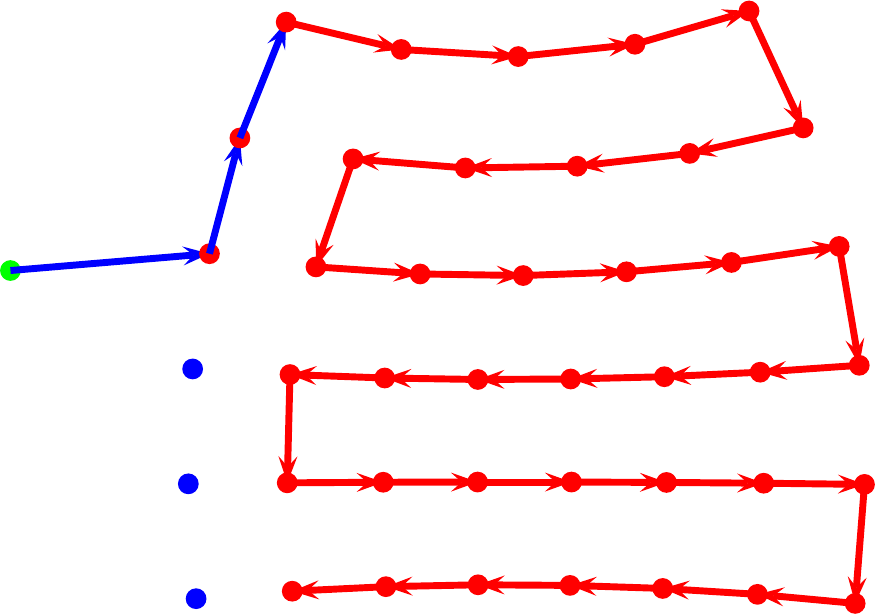}
    \label{fig:rti_even_4}
  }
  \subfloat[]{
    \includegraphics[width=0.28\columnwidth]{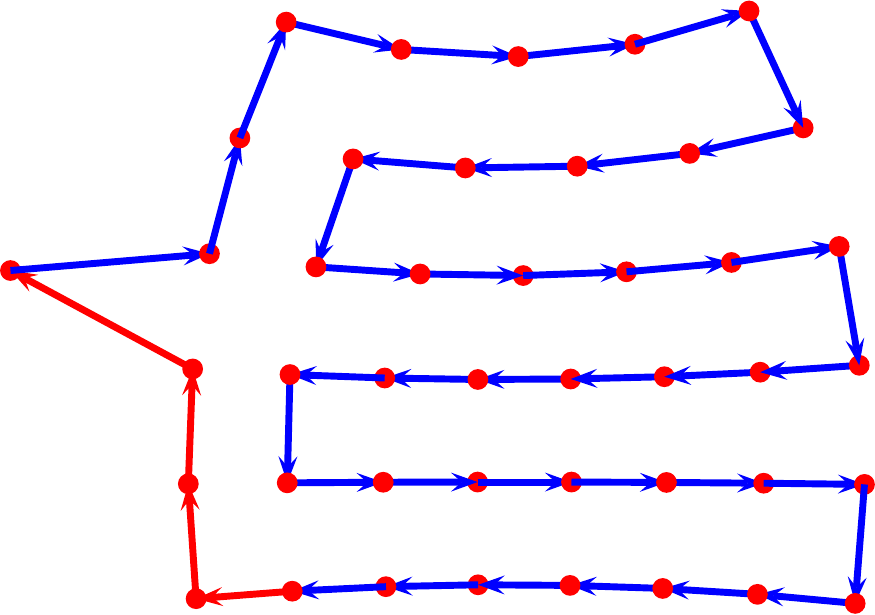}
    \label{fig:rti_even_6}
  }
  \vspace{-0.2cm}
  \caption[Illustration of the procedure of determining the predictable sequence of RTI positions for even number of horizontal rows.]{Illustration of the procedure of determining the safety pilot predictable sequence of RTI positions for even number of horizontal rows. The green dot marks the initial state $P_{i}$, the blue dots stand for the unvisited RTI positions, and the red dots for already visited RTI positions. The arrows show the transitions between particular RTI positions, where the red arrows stand for the transitions added during the last step.}
  \label{fig:rti_even_tutorial}
  \vspace{-0.4cm}
\end{figure}

In the case of an even number of horizontal rows, the RTI positions are added line by line with switching the left-right and the right-left direction, and omitting the points that lie on the same vertical boundary as $P_{s}$. After reaching the last admissible position in the most bottom row, the remaining points are added from the bottom row up to and including the $P_{e}$ into $S_p$. Finally, the $P_{i}$ is added at the end of the $S_p$ to ensure the return to the initial position.
The graphical illustration of this process is shown in~\cref{fig:rti_even_3,,fig:rti_even_4,fig:rti_even_6}.

The solution for an odd number of horizontal lines is derived from the solution for even number of rows with several modifications. Firstly, the pair of consequent horizontal lines (indicated by pair of indices $(h_{o, 1}, h_{o, 2})$) with minimum number of RTI positions is determined.
Then, the solution for an odd number of rows is the same as in the case of even number of rows until the procedure reaches the pair $(h_{o, 1}, h_{o, 2})$. The particular RTI positions within this pair of rows are traversed either in an up-side-down-side manner ~(see~\autoref{fig:rti_comparison_mine} for example), or by following the positions in row $h_{o, 1}$ to the opposite side, then flying back to the starting side of row $h_{o, 2}$ and again following this row to the opposite side. From there, the situation and also the solution is again the same as in the case of an even number of rows.

\subsection{Trajectory Generation and Tracking}
The desired trajectory for the RTI ($\Omega$) is generated by the sampling of direct straight paths between consequent RTI positions with the sampling distance $d_{RTI}$ (computed based on the desired velocity), without considering any obstacles.
To achieve precise lighting conditions, the UAV carrying the light is supposed to hover at the desired position while taking a photo. This requirement is introduced into the presented system by multiple recurrences of the desired RTI position as the transition point of $\Omega$ after each fly-over to the next RTI position. The number of these repetitions is proportional to the time required for the stabilization.

To achieve a reliable following of the desired trajectory $\Omega$ in an environment with obstacles and in the presence of disturbances, which cannot be omitted in real systems, the trajectory tracking during the RTI procedure is defined as an optimization task within the MPC framework in the proposed system.
Thanks to the independence of the position and orientation control in case of multi-rotor UAVs, the optimization loop can be divided into two separate tasks.

The position control is defined as a nonlinear constrained optimization task over a sequence of control inputs $\mathcal{U}_p(t)$ starting at time $t$ with an objective function $J_p$, and set of nonlinear constraints $g_p(\cdot)$ on the horizon of length $N$ as
\begin{equation}
  \begin{aligned}
    \mathcal{U}_p(t)^\ast = \arg &\min J_p(\mathcal{U}_p(t)), \, \, \\
                                 &s.\, t.\, \, g_p(\mathcal{U}_p(t), \mathcal{O}(t)) &\leq 0,
  \end{aligned}
\end{equation}
where $\mathcal{O}(t)$ is the set of all obstacles present at time $t$ in the environment, including the UAV carrying the camera.

The objective function $J_p(\cdot)$ is defined as the weighted sum
\begin{equation}\label{eq:objective_function_position}
  J_p = \alpha J_{pos} + \beta J_c + \gamma J_{obs} + \delta J_{rti},
\end{equation}
where $J_{pos}$ stands for the part penalizing the deviations from the desired trajectory, $J_c$ is the part penalizing the changes in sequence of control inputs, and $J_{obs}$ responds for the penalization of trajectories in the proximity of obstacles. The value of $J_p$ is increased by adding $J_{rti}$ for trajectories that lead to occlusions caused by the UAV carrying light or lead to shades in the image caused by the lighting from behind the UAV carrying the camera. Coefficients $\alpha, \beta, \gamma,\,\text{and}\,\delta$ are weights used for the scaling of particular parts of the objective function.

\begin{figure} [h]
  \centering
  \subfloat[top view]{
    \includegraphics[width=0.45\columnwidth]{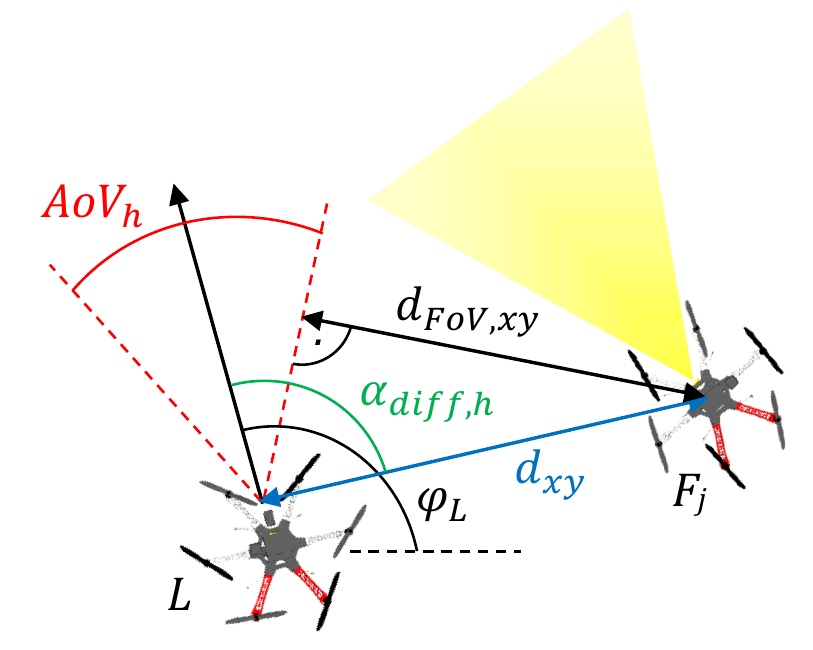}
    \label{fig:top_view}
  }
  \subfloat[side view]{
    \includegraphics[width=0.45\columnwidth]{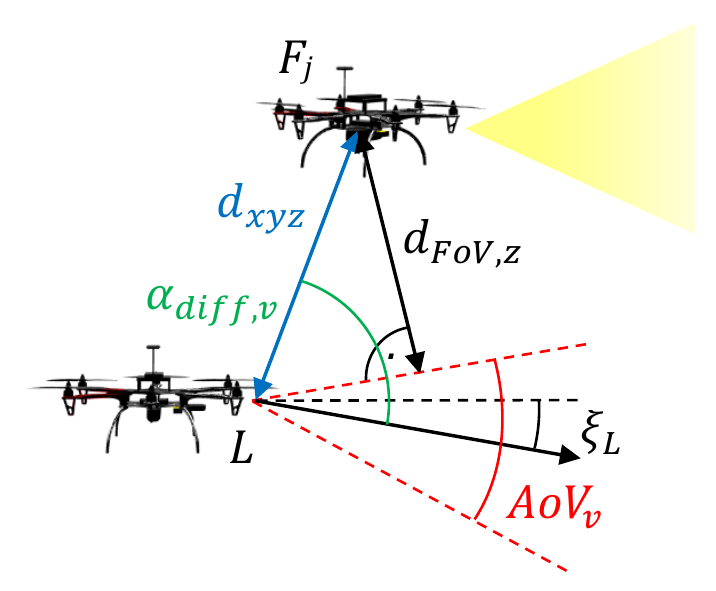}
    \label{fig:side_view}
  }
  \vspace{-0.1cm}
  \caption{Graphical illustration of meaning of particular symbols used in equations~\eqref{eq:rti_occlusion} for computation of part of the objective function penalizing the occlusion caused by the UAV carrying a light}
  \vspace{-0.4cm}
  \label{fig:occlusion_explanation}
\end{figure}

The function $J_{rti}$, which was proposed specifically for this application, is defined as
\begin{equation}\label{eq:rti}
  J_{rti} = \sum_{k=1}^N \left(\min\left\{0, \frac{d_{FoV}(k) - r_{d, FoV}}{d_{FoV}(k) - r_{a, FoV}}\right\}\right)^2,
\end{equation}
where $r_{d, FoV}$ and $r_{a, FoV}$ are detection and avoidance radii with respect to camera Field of View (FoV), and $d_{FoV}(\cdot)$ stands for the distance from the nearest border of the FoV. This distance can be computed according to equations
\begin{align}\label{eq:rti_occlusion}
  d_{xy}(k) &= \sqrt{(x_c(k)-x_l(k))^2 + (y_c(k)-y_l(k))^2},\nonumber\\
  \beta_{diff,h}(k) &= \min\{\alpha_{diff, h}(k), \pi - \alpha_{diff, h}(k)\},\nonumber\\
  \beta_{diff,v}(k) &= \min\{\alpha_{diff, v}(k), \pi - \alpha_{diff, v}(k)\},\nonumber\\
  d_{FoV, xy}(k) &= d_{xy}(k) \sin\left(\beta_{diff, h}(k) - \frac{AoV_h}{2}\right),\\
  d_{FoV, z}(k) &= \text{dist}(P_l(k), P_c(k)) \sin\left(\beta_{diff, v}(k) - \frac{AoV_v}{2}\right),\nonumber\\
  d_{FoV}(k) &= \sqrt{d_{FoV, z}(k)^2 + d_{FoV, xy}(k)^2} - r_d,\nonumber
\end{align}
where $P_l(k) = [x_l, y_l, z_l]^T$ and $P_c(k) = [x_c, y_c, z_c]^T$ is the position of  UAV carrying light and UAV carrying camera at the time corresponding to $k$-th transition point. $AoV_h$ and $AoV_v$ are horizontal and vertical angles of the camera FoV, $d_{FoV, xy}(k)$ is the distance to the nearest vertical border of FoV, $d_{FoV, z}(k)$ is the distance to the nearest horizontal border of FoV, and $r_d$ marks the radius of the UAV. $\alpha_{diff,h}(k)$ and $\alpha_{diff,v}(k)$ stand for the angle between the nearest vertical respectively horizontal border of the FoV and connecting line between UAV carrying camera and UAV providing light. $\beta_{diff,h}(k)$ and $\beta_{diff,v}(k)$ are equivalent to $\alpha_{diff,h}(k)$ and $\alpha_{diff, v}(k)$, but besides the FoV of the camera, they include also the FoV of the virtual camera pointed in the exact opposite direction than the real camera. With this alteration, the $J_{rti}$ penalizes not only the occlusion caused by the UAV carrying the light but also the shadows visible in the FoV caused by lighting from behind the UAV carrying the camera, which is important for the RTI image processing.
The graphical illustration of symbols used in equation~\eqref{eq:rti_occlusion} is shown in~\autoref{fig:occlusion_explanation}.

The set of nonlinear constraints $g_p(\cdot) \leq 0$ can be broken down into the following constraints
\begin{equation}
  \begin{split}
    g_c(\mathcal{U}_p(k)) &\leq 0, \forall k \in \{1, ..., N\},\\
    g_{obs}(P_l(k), \mathcal{O}(t)) &\leq 0, \forall k \in \{1, ..., N\},\\
    g_{rti}(P_{l}(k), \mathcal{O}(t), \psi_c(k)) &\leq 0, \forall k \in \{1, ..., N\},
  \end{split}
\end{equation}
where $\psi_c(k)$ stands for the configuration of the UAV carrying camera, $g_c(\cdot)$ includes the limitations on control inputs, $g_{obs}(\cdot)$ defines the infeasibility of trajectories colliding with obstacles, and $g_{rti}(\cdot)$ complements the objective function $J_{rti}$ by defining the entire FoV as an unfeasible region.
{\color{black}

  In a similar manner, the process of finding the optimal sequence of orientation control inputs $\mathcal{U}_o(t)$ on the horizon of length $N$ can be defined as the quadratic constrained optimization task with the objective function $J_o(\cdot)$ and set of nonlinear constraints $g_o(\cdot)$ as
  \begin{equation}
    \begin{split}
      \mathcal{U}_o(t)^\ast &= \arg \min J_o(\mathcal{U}_o(t)),\\
                            &s. \, \, t.\, \, g_o(u_o(k),O_{j}(k)) \leq 0, \forall k \in \{1, ..., N\}.
    \end{split}
  \end{equation}

  The objective function $J_o(\cdot)$ consists of two parts
  \begin{equation}
    J_o = \zeta J_{or} + \kappa J_{co},
  \end{equation}
  where $J_{or}$ is the part penalizing the deviation from the desired orientation of light, $J_{co}$ stands for the part penalizing the fast changes in consequent control inputs $u_{o}(\cdot)$, and $\zeta$ and $\kappa$ are weights used for scaling of parts of the objective function $J_o(\cdot)$. The set of nonlinear constraints $g_o(\cdot) \leq 0,$
  $\forall k \in \{1, ..., N\}$ can be split into the following constraints
  \begin{equation}
    \begin{split}
      g_{co}(u_o(k)) &\leq 0, \forall k \in \{1, ..., N\},\\
      g_{or}(O_{l}(k)) &\leq 0, \forall k \in \{1, ..., N\},
    \end{split}
  \end{equation}
  where $O_l(\cdot)$ is the orientation of the light carried by the UAV, $g_{co}(\cdot)$ stands for the constraints introducing the limits on control inputs, and $g_{or}(\cdot)$ introduces the limitations on angles that define the orientation of the light.

%% file: chapters/experimental_results.tex
\section{EXPERIMENTAL RESULTS}\label{sec:experimental_results}

\subsection{Performance of generation of lighting positions sequence}
The purpose of this section is to qualitatively and quantitatively compare algorithm SPPA, FIB-LKH and their combination which applies SPPA part for the generation of the desired lighting positions and LKH solver for finding a path connecting these positions (further referenced as LKH).
The test was performed on the testing case of 10000 samples, each with randomly chosen parameters $\lambda_{v, min}, \lambda_{v, max}, \lambda_{h, min}, \lambda_{h, max}, d_l, v_s$, and initial position of the UAV carrying the light $P_{i}$.

The quality of solutions was compared regarding time requirements and the length of the resulting path. Concerning the CPU time, the SPPA is faster than the others. However, since the total CPU time needed by any method does not exceed \SI{0.5}{s} for all considered problems (computed on the single-core CPU Intel CORE i7 8250), this aspect is not important for our application. More significant parameter is the length of the paths produced by particular methods. Considering this criterion as the comparison value, SPPA is better or equals to LKH solution in $9\%$ of test samples and is not longer by more than $50\%$ in $98\%$ of test samples. Paths generated by FIB-LKH approach are mostly the shortest among all methods. However, they do not fully exploit the borders of the defined scanning area (see~\autoref{fig:rti_comparison_fib}). More detailed results of the quantitative comparison are shown in~\autoref{fig:rti_quantiles}.
Examples of generated sequences by particular methods used for qualitative comparison are shown in~\autoref{fig:rti_comparison}.
\begin{figure}[h]
  \vspace{-0.1cm}
  \centering
  \input{fig/rti_quantiles.tex}
  \vspace{-0.2cm}
  \caption{Comparison of length of paths obtained by SPPA, LKH, and Fib-LKH approach}
  \vspace{-0.9cm}
  \label{fig:rti_quantiles}
\end{figure}
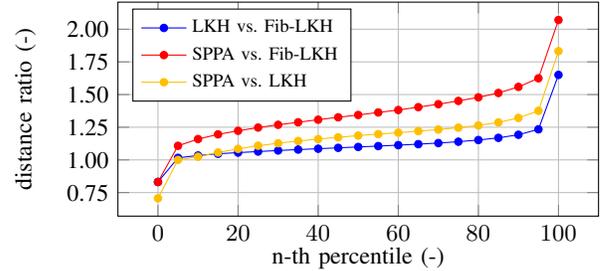
\begin{figure}[h]
  \centering
  \subfloat[]{
    \includegraphics[width=0.27\columnwidth]{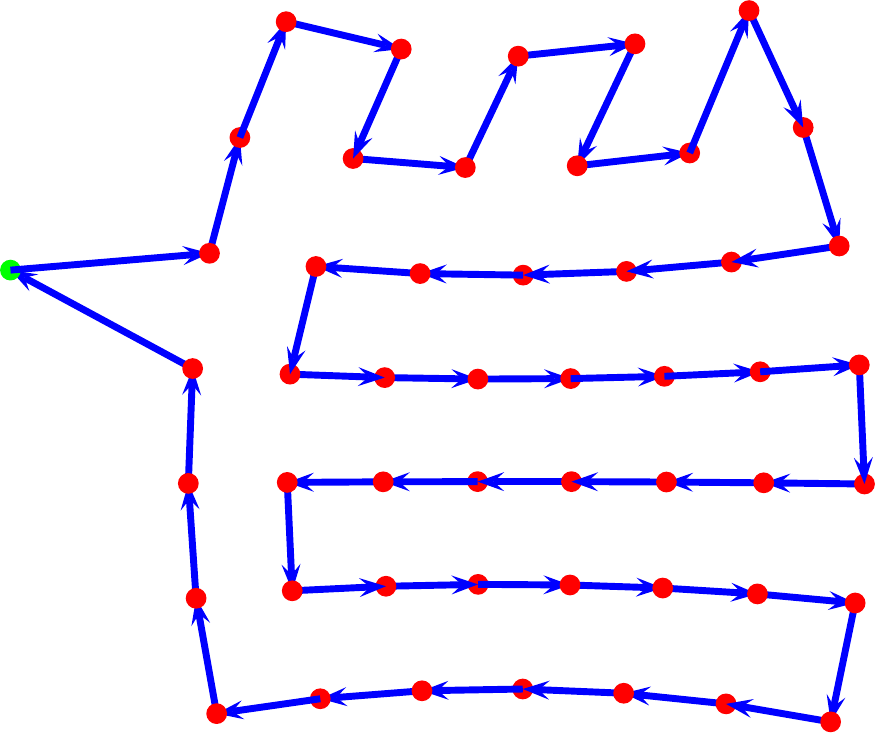}
    \label{fig:rti_comparison_mine}
  }
  \subfloat[]{
    \includegraphics[width=0.27\columnwidth]{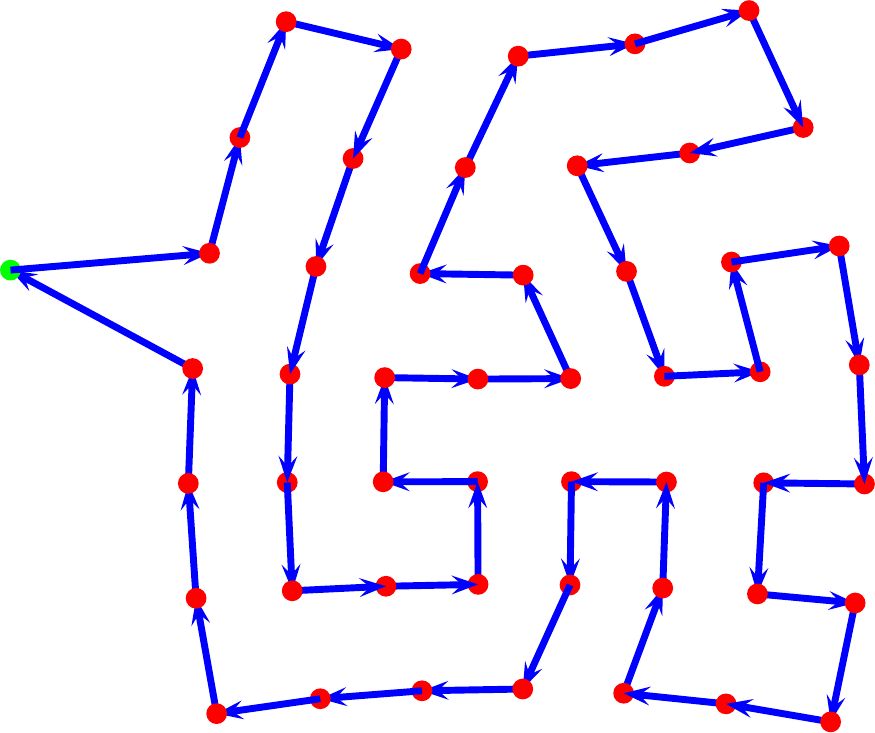}
    \label{fig:rti_comparison_lkh}
  }
  \subfloat[]{
    \includegraphics[width=0.27\columnwidth]{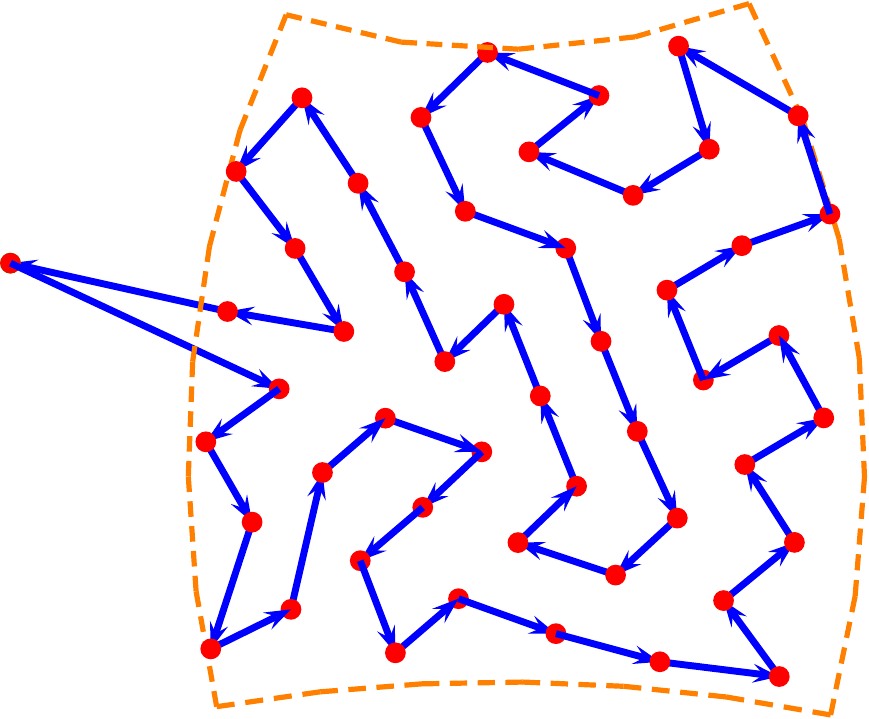}
    \label{fig:rti_comparison_fib}
  }
  \vspace{-0.2cm}
  \caption[]{Comparison of the solution obtained with SPPA (a), the solution generated by LKH method (b), and the solution generated by Fib-LKH approach (c). The orange dashed line marks the borders of the defined scanning area.}
  \vspace{-0.4cm}
  \label{fig:rti_comparison}
\end{figure}

\subsection{Verification of the overall RTI approach}
The deployability of the SPPA method for the realization of RTI scanning, which was chosen together with experts from the Czech National Heritage Institute, who are potential end users of the proposed system, as the best variant for its real deployment, was verified through several experiments in the realistic robotic simulator Gazebo and within real-world experiments deploying two autonomous UAVs in the interior of the Church of St. Mary Magdalene in Chlum\'{i}n.

\begin{figure}[h]
  \centering
  \subfloat[$xy$-view]{
    \includegraphics[width=0.47\columnwidth]{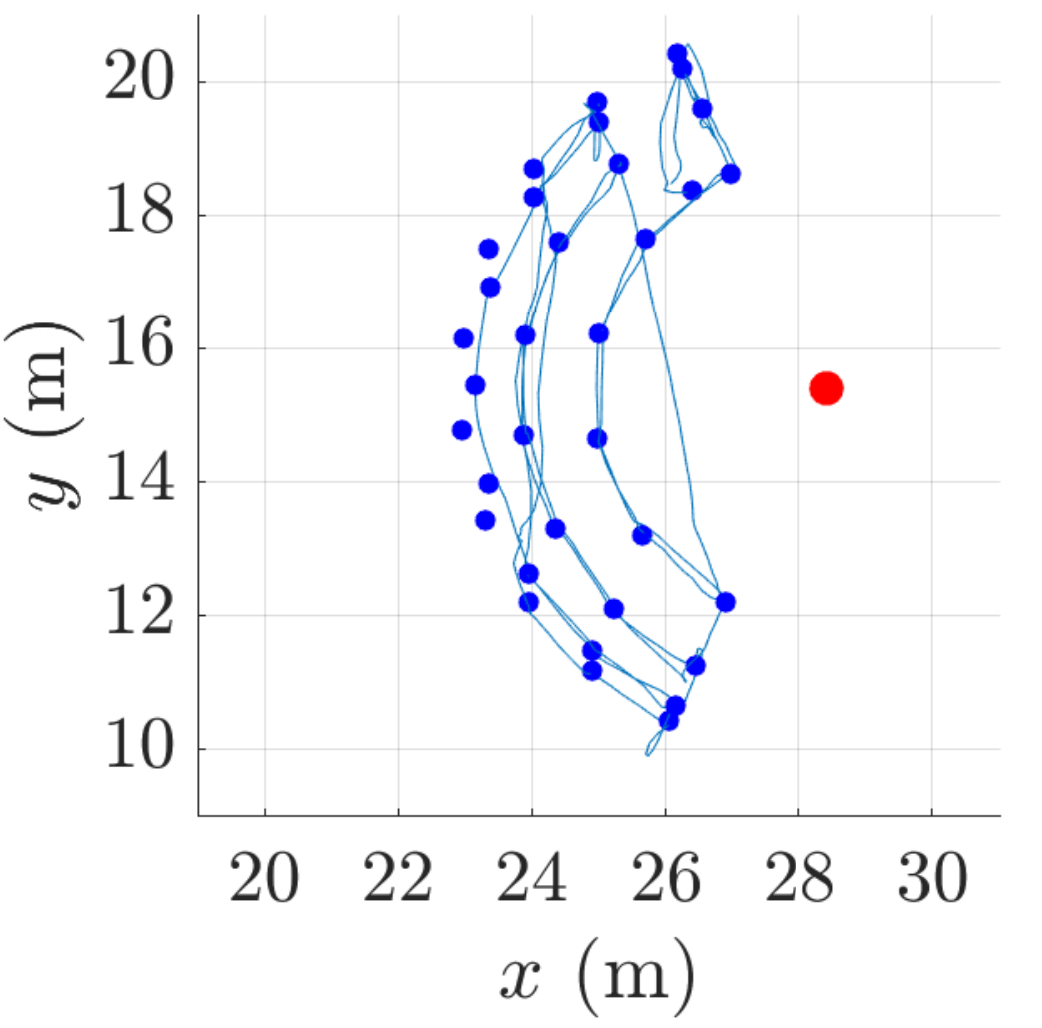}
    \label{fig:rti_exp_trajectory_3}
  }
  \subfloat[$yz$-view]{
    \includegraphics[width=0.47\columnwidth]{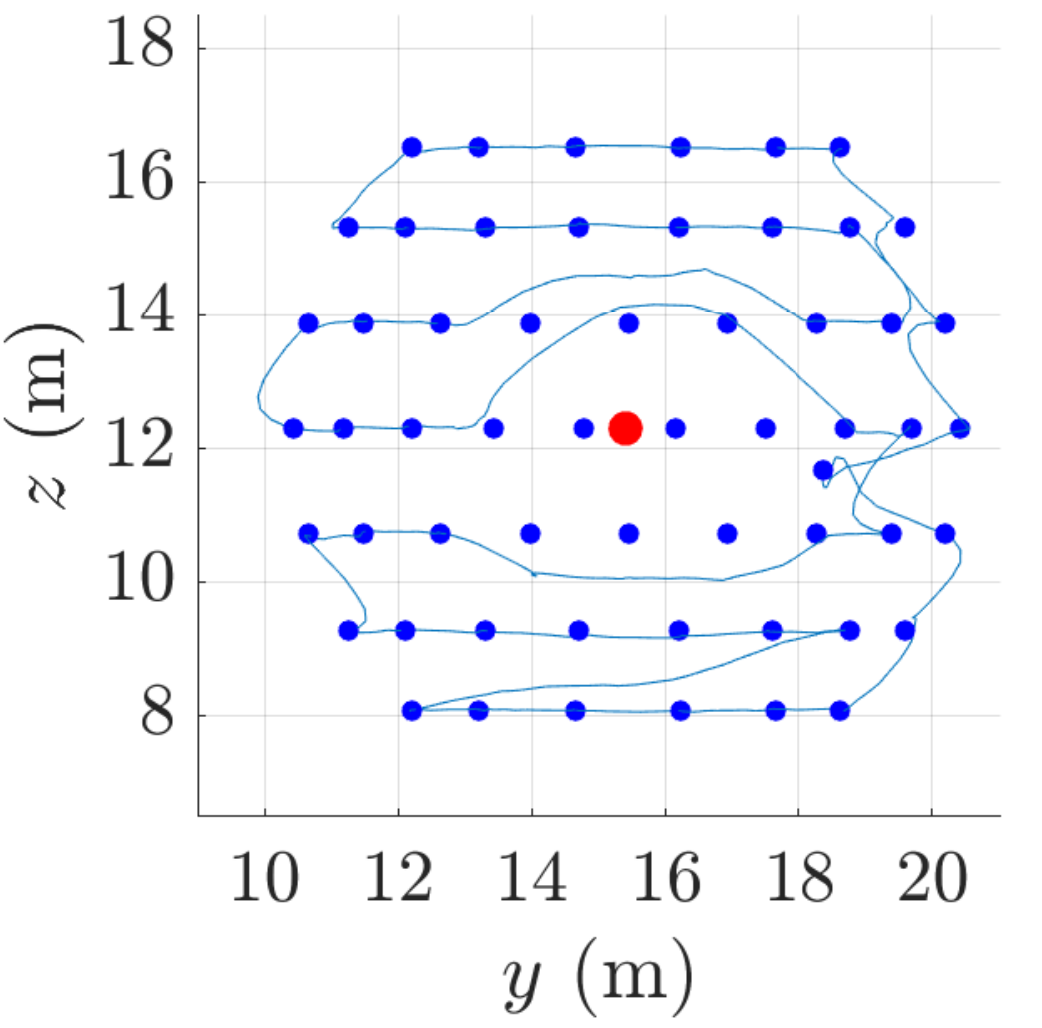}
    \label{fig:rti_exp_trajectory_4}
  }
  \vspace{-0.3cm}
  \hfill
  \subfloat[yaw - light]{
    \resizebox{0.45\linewidth}{!}{
      \input{fig/exp_rti_follower_yaw.tex}
    }
    \label{fig:rti_exp_trajectory_yaw}
  }
  \subfloat[pitch - light]{
    \resizebox{0.45\linewidth}{!}{
      \input{fig/exp_rti_follower_pitch.tex}
    }
    \label{fig:rti_trajectory_pitch}
  }
  \caption[The generated RTI positions and the trajectory flown by the UAV carrying the light during the RTI scanning procedure.]{The generated RTI positions and the trajectory flown by the UAV carrying the light during the RTI scanning procedure. The blue dots mark particular RTI positions, the red dot marks the position of the scanned object, and the blue line shows the trajectory.}
  \label{fig:rti_exp_trajectories}
  \vspace{-0.5cm}
\end{figure}

The presented simulation in which the RTI scanning procedure is performed on the statue situated above the altar leads to the generation of 56 RTI positions and the resulting trajectory of the overall length \SI{110.55}{m}.
The set of generated points together with the trajectory flown by the UAV carrying the light are shown in~\cref{fig:rti_exp_trajectories}.
In compliance with the theory presented in~\autoref{sec:rti_scanning}, the UAV carrying light stops at each reachable RTI position and waits until an image is taken by the UAV carrying the camera. In this way, the system collects 56 images of the scanned object under various lighting conditions. The images are then registered to each other to compensate for the motion of UAV carrying a camera during the scanning process. Based on the registered images and the file containing the information about corresponding lighting directions, the PTM representation of the image is computed with the use of program PTM Fitter.

The main advantage of obtaining the PTM from the set of images is that the image can be displayed under arbitrary lighting conditions. Since this result can be hardly presented within the printed work, the resulting PTM representation of the scanned object, obtained from the images taken by an onboard camera, is shown in the video available at \url{http://mrs.felk.cvut.cz/papers/rti2020ral}.

\begin{figure}[h]
  \vspace{-0.5cm}
  \centering
  \hspace*{-0.6em}
  \subfloat[upper-left light]{
    \includegraphics[width=0.325\columnwidth]{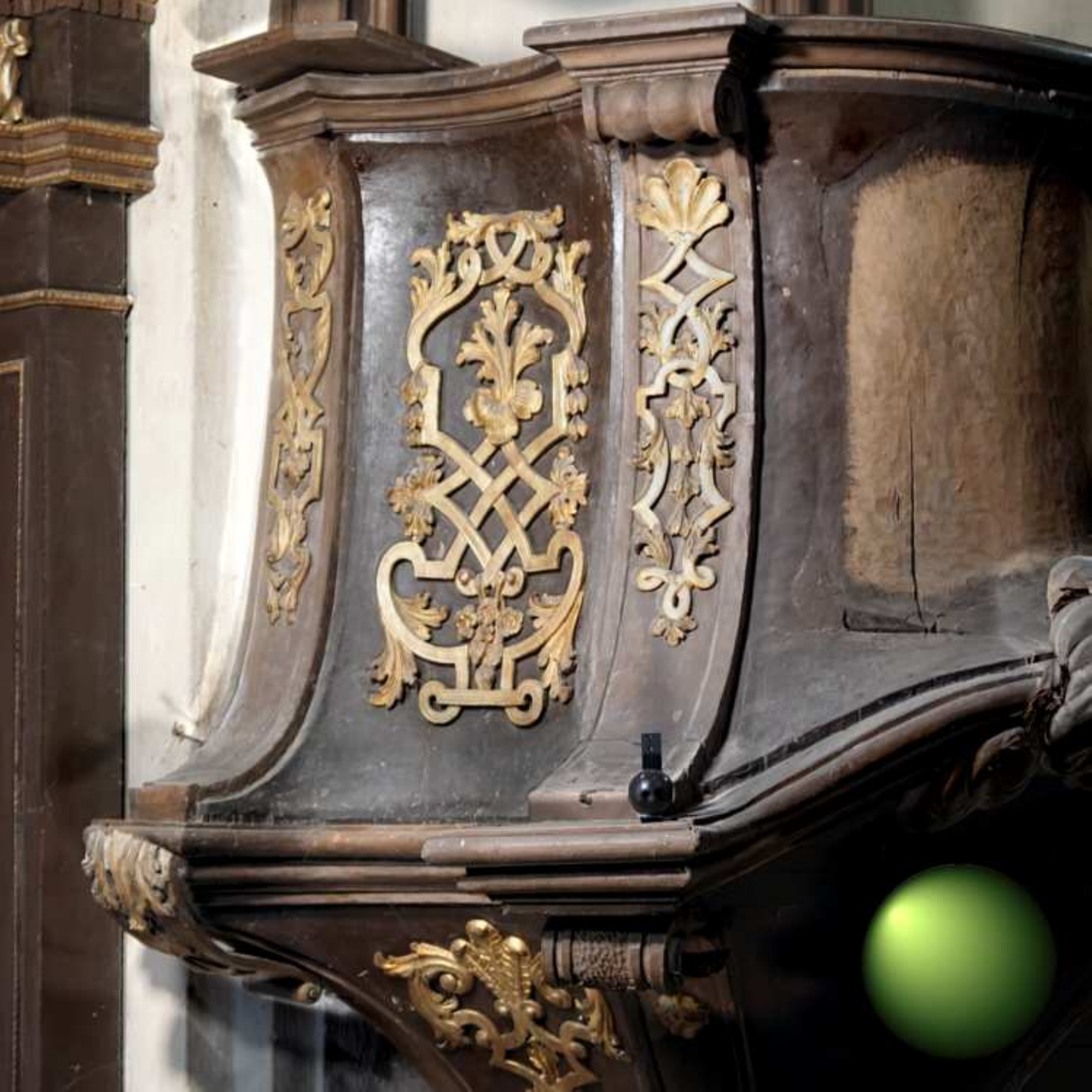}
    \label{fig:ptm_up_left}
  }
  \hspace*{-0.9em}
  \subfloat[left light]{
    \includegraphics[width=0.325\columnwidth]{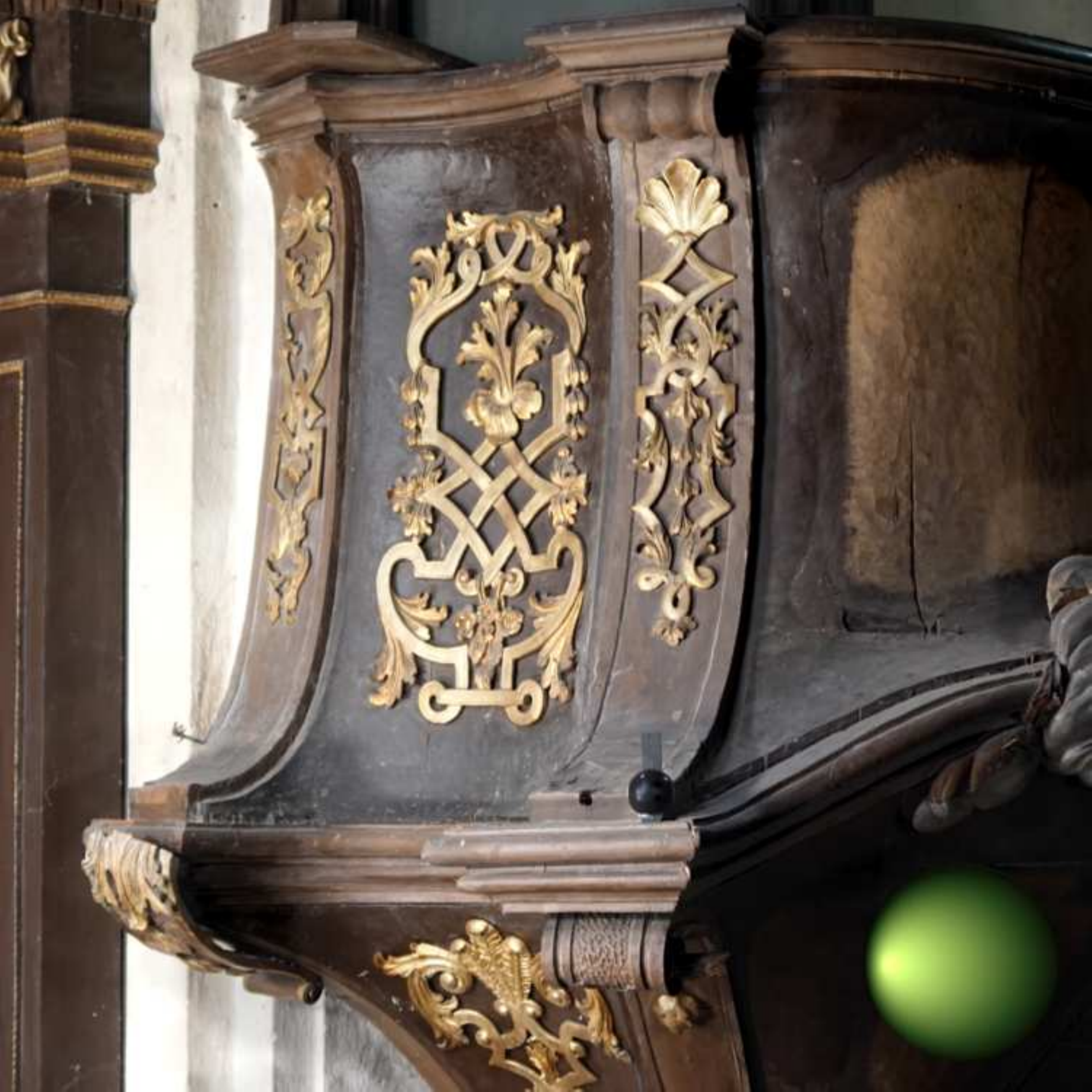}
    \label{fig:ptm_left_90}
  }
  \hspace*{-0.9em}
  \subfloat[bottom-right light]{
    \includegraphics[width=0.325\columnwidth]{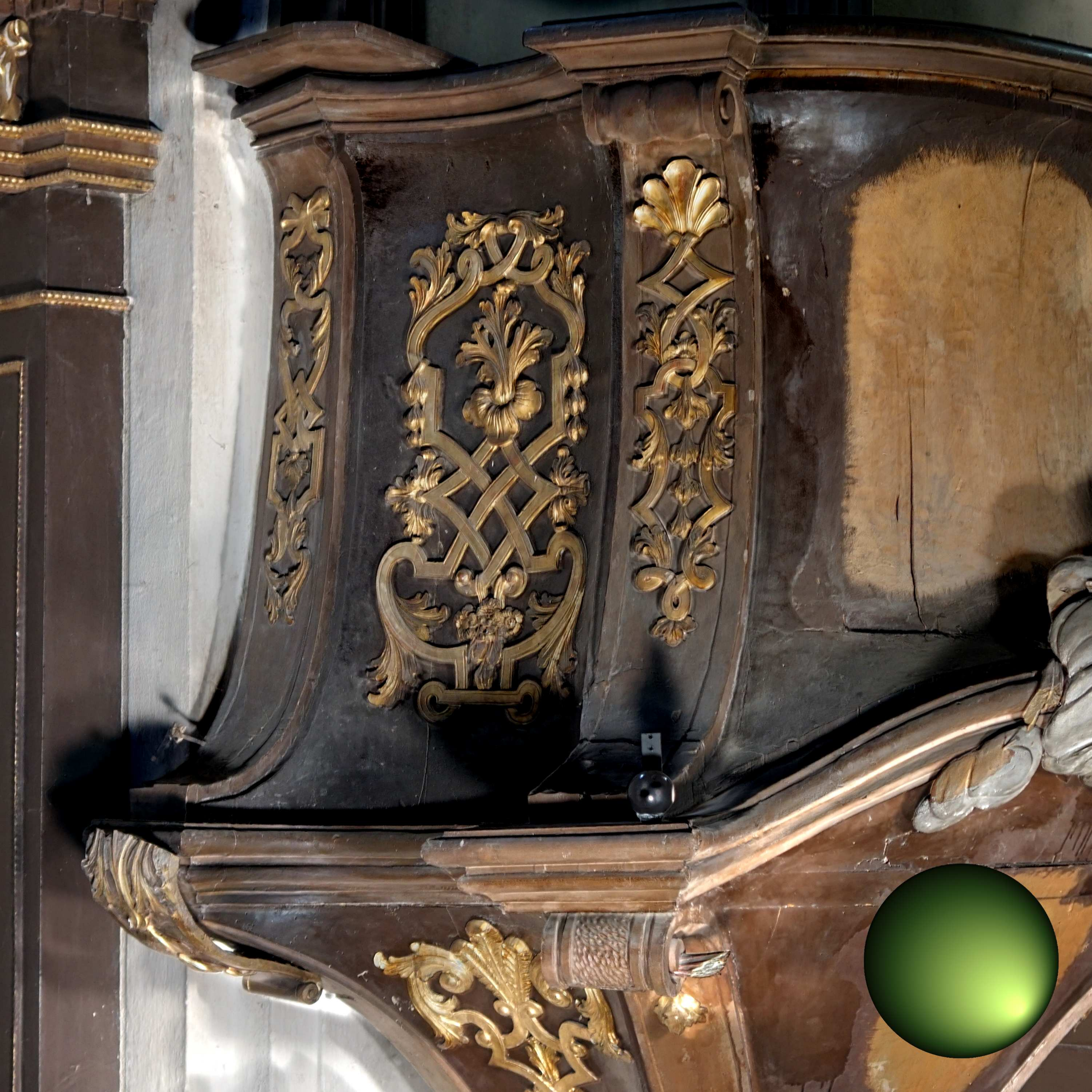}
    \label{bottom-right light}
  }
  \vspace{-0.1cm}
  \caption[Presentation of the PTM representation of the scanned object obtained from images taken by UAV during the RTI experiment performed in Church of St. Mary Magdalene in Chlum\'{i}n.]{Presentation of the PTM representation of the scanned object obtained from images taken by UAV during the RTI experiment performed in Church of St. Mary Magdalene in Chlum\'{i}n. For video see~\hbox{\url{http://mrs.felk.cvut.cz/papers/rti2020ral}}.}
  \vspace{-0.1cm}
  \label{fig:rti_ptm_results}
\end{figure}

The real experiment was adapted to fit into the restricted space of the church in Chlum\'{i}n. To enable the comparison of results of the proposed method and H-RTI, the object of interest (part of the pulpit) was chosen in the height accessible by people and it was illuminated from the same 12 positions \hbox{(see~\autoref{fig:rti_rviz})} by two different approaches - with the camera carried by an autonomous UAV \hbox{(see~\autoref{fig:real_world_deployment})} and with the camera mounted on a static tripod~\hbox{(see~\autoref{fig:experiment_tripod})}. The latter approach eliminates the imprecision caused by the camera motion and hence enables the objective comparison of the results obtained from the same set of images with lighting vectors computed from the reflections on the black ball (H-RTI) and from the position of the light-carrying UAV provided by the application-tailored localization system\hbox{~{\cite{icra20petracek}}}. The images generated based on the PTM representation of the scanned object are shown in~\mbox{\autoref{fig:rti_ptm_results}} and in the video available at \mbox{\url{http://mrs.felk.cvut.cz/papers/rti2020ral}}.

\subsection{Dependence of PTM quality on precision of localization}
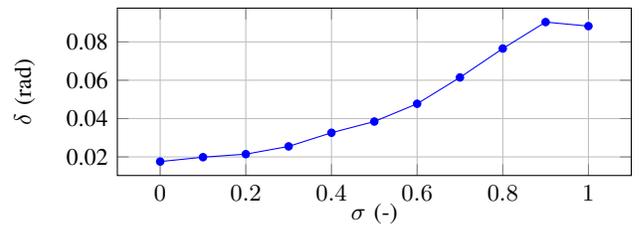
\begin{figure}[h]
  \centering
  \input{fig/normals_quantitative_comparison.tex}
  \vspace{-0.1cm}
  \caption{The dependence of the average error in normals $\delta$ on the simulated localization error represented by $\mathcal{N}(0, \sigma^2)$ for positional error (m) and $\mathcal{N}(0, (\frac{2\pi\sigma}{36})^2)$ for orientation error (rad). Values of $\delta$ for a particular $\sigma$ is computed as an average result of 20 experiments.}
  \vspace{-0.6cm}
  \label{fig:graph_accuracy_comparison}
\end{figure}

\begin{figure}[h]
  \hspace*{-0.6em}
  \centering
  \subfloat[]{
    \includegraphics[width=0.29\columnwidth]{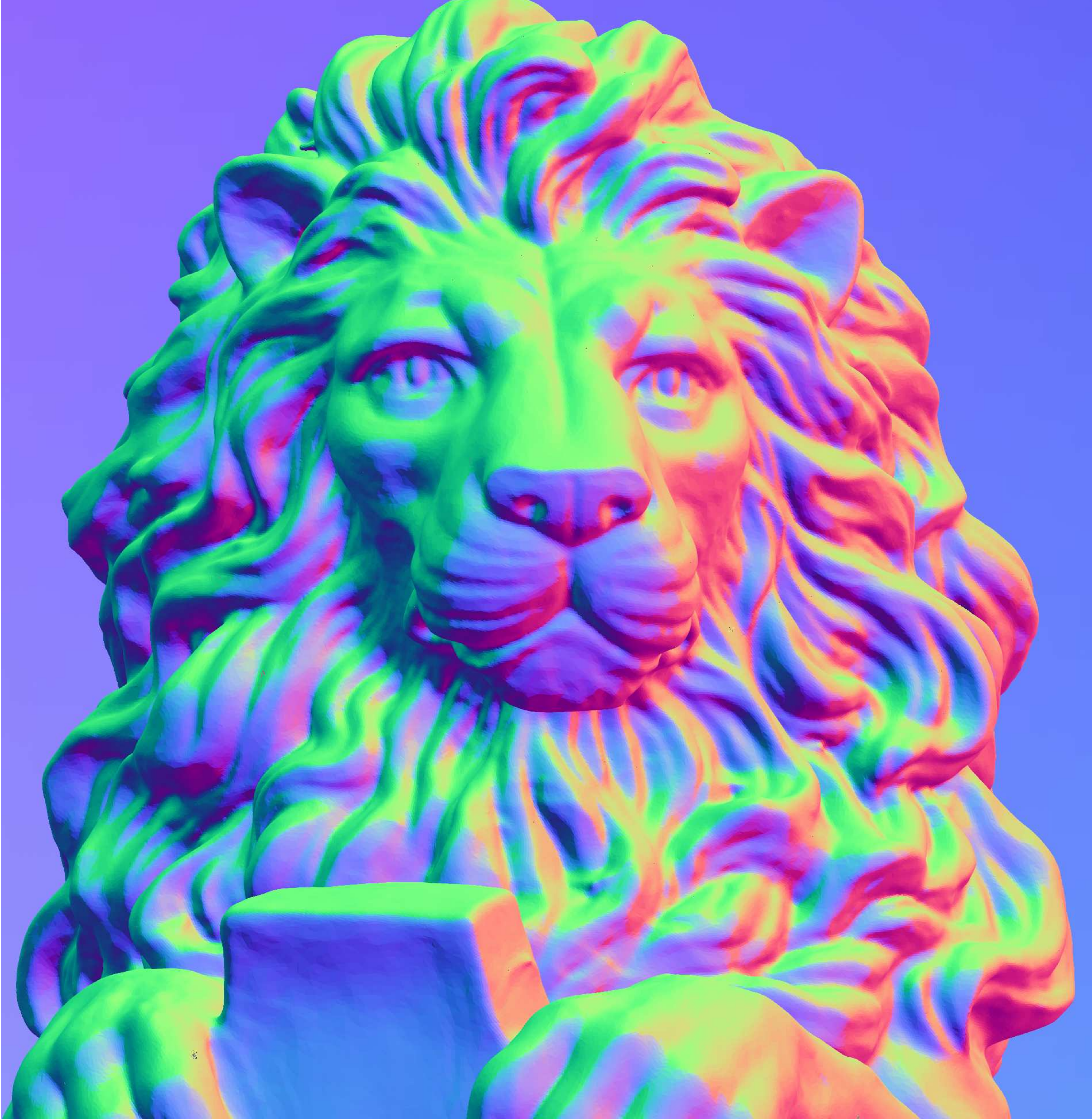}
    \label{fig:rti_lion_gt}
  }
  \hspace*{-1.0em}
  \subfloat[]{
    \includegraphics[width=0.29\columnwidth]{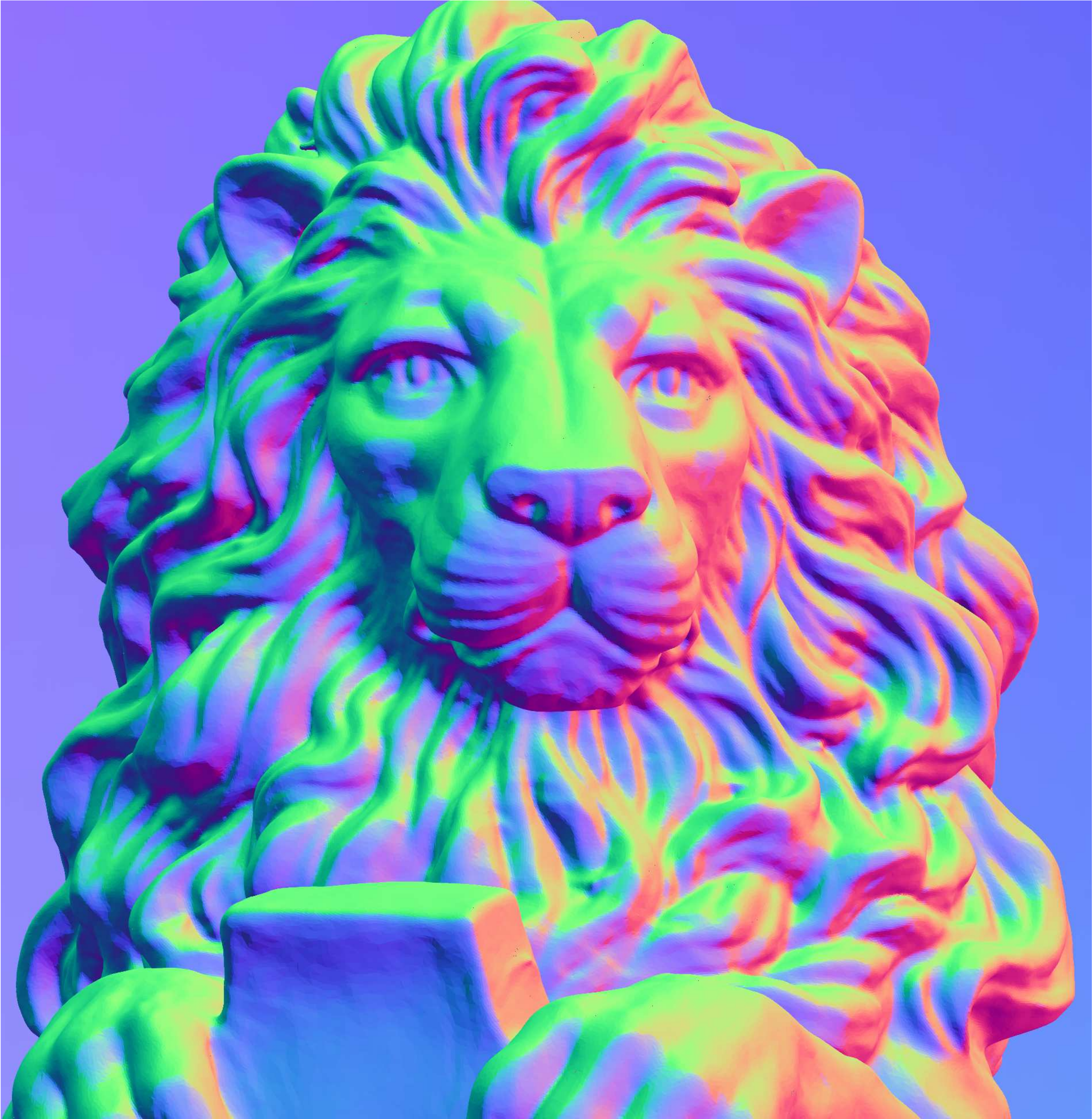}
    \label{fig:rti_lion_stddev}
  }
  \hspace*{-1.0em}
  \subfloat[]{
    \includegraphics[width=0.40\columnwidth]{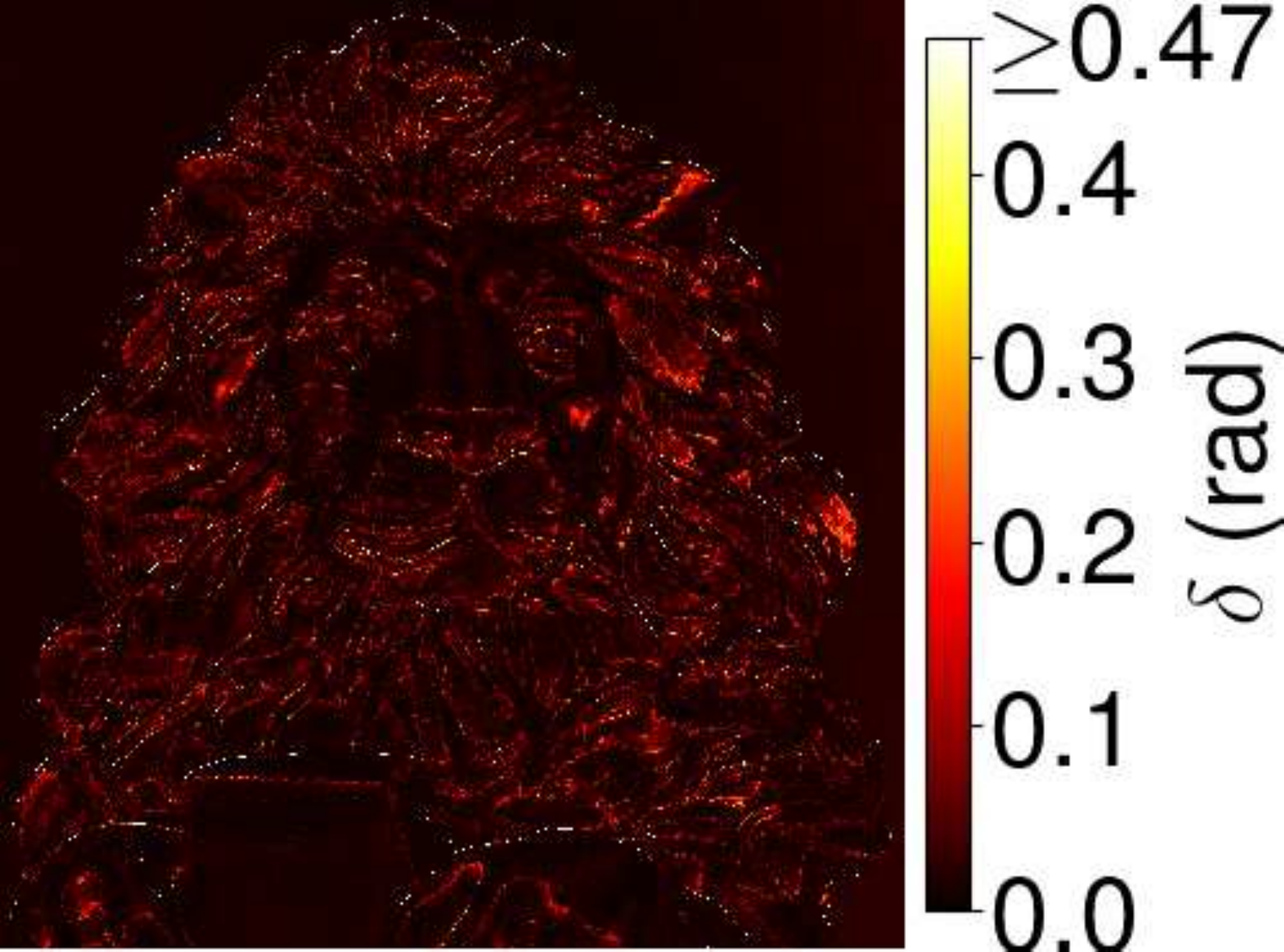}
    \label{fig:rti_lion_heatmap}
  }
  \caption[]{Comparison of the normal map obtained with SPPA (360 positions and zero localization error) (a) used as a ground truth and normal map obtained with SPPA (60 positions and localization error (m) modeled by $\mathcal{N}(0,\,0.09)$ for position and $\mathcal{N}(0,\,0.003)$ for orientation) (b). Figure (c) shows the size of angle between normal vectors in maps (a) and (b) for particular pixels.}
  \vspace{-0.4cm}
  \label{fig:rti_lion_comparison}
\end{figure}

To examine how the precision of localization affects the quality of the resulting PTM, a simulation-based quantitative comparison was conducted. The whole RTI procedure was performed on a lion statue with localization error sampled from the normal distribution with zero mean and multiple distinct values of standard deviation. The normal map obtained using SPPA (60 positions) and a modelled localization error is compared to the normal map obtained with SPPA (360 positions and zero localization error) used as ground truth. The results of this comparison are presented in~\hbox{\autoref{fig:graph_accuracy_comparison}}. The average difference from the ground truth normals for the normal map obtained for the precision of localization presented in\hbox{\cite{icra20petracek}} is \SI{0.026}{rad}~(see~\hbox{\autoref{fig:rti_lion_comparison}} for details). This value is lower than the average difference caused by the misplacement of the reflective ball with respect to the center of the scanned object within the \hbox{H-RTI} procedure, which is unavoidable in this method.


\subsection{Comparison of the proposed approach with H-RTI}
\begin{figure}[h]
  \centering
  \hspace*{-0.6em}
  \subfloat[]{
    \includegraphics[width=0.29\columnwidth]{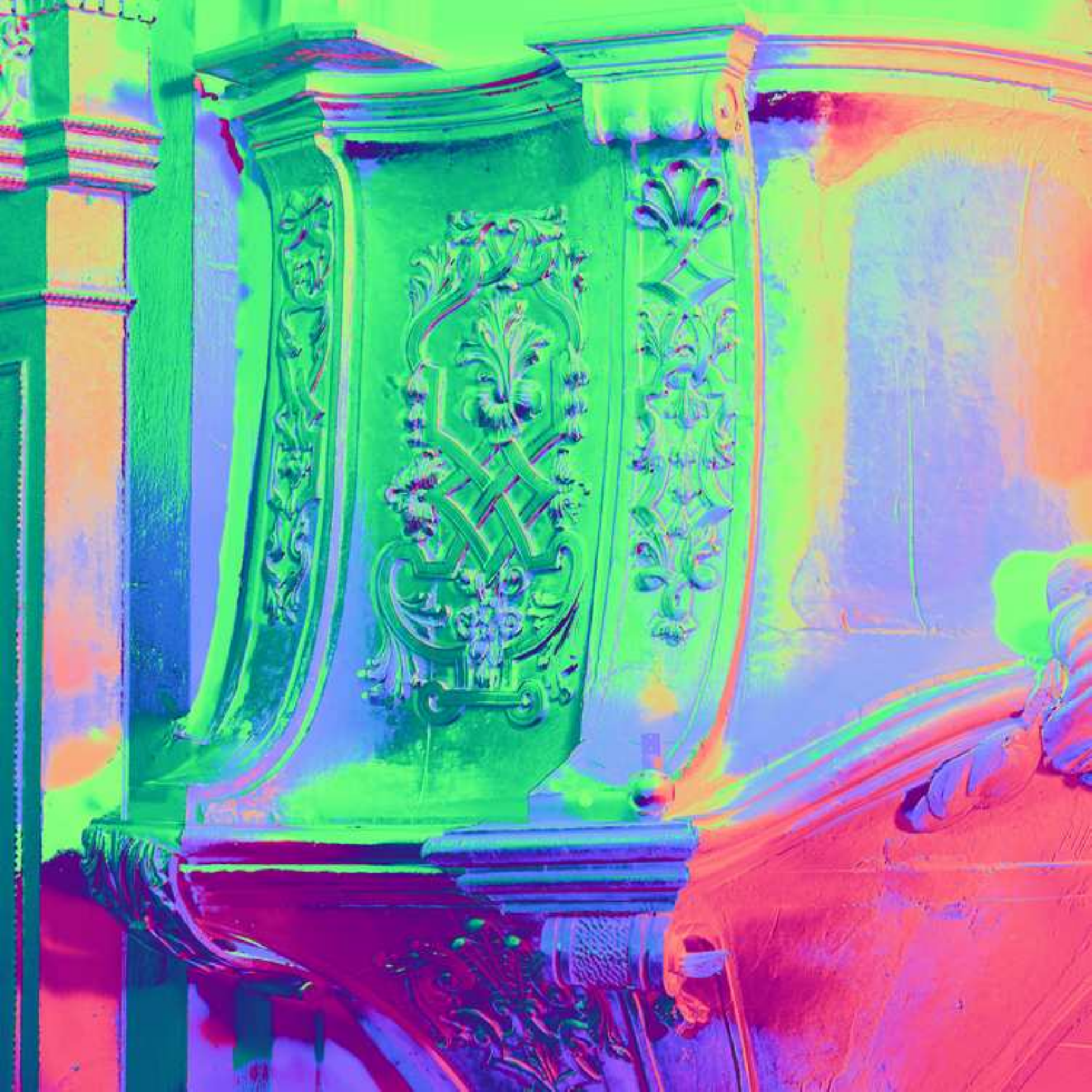}
    \label{fig:hrti_normal_map}
  }
  \hspace*{-0.9em}
  \subfloat[]{
    \includegraphics[width=0.29\columnwidth]{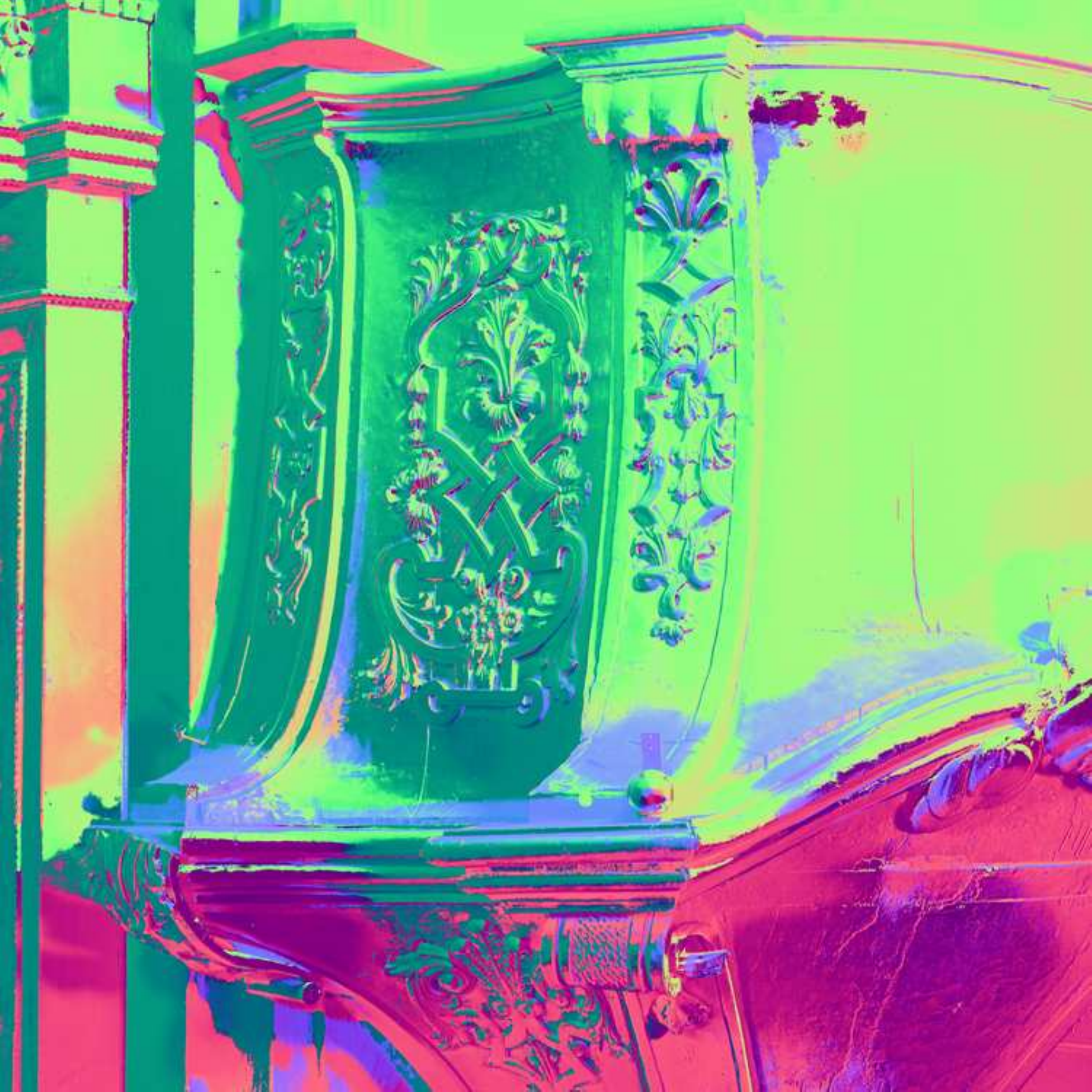}
    \label{fig:mine_rti_normal_map}
  }
  \hspace*{-0.9em}
  \subfloat[]{
    \includegraphics[width=0.388\columnwidth]{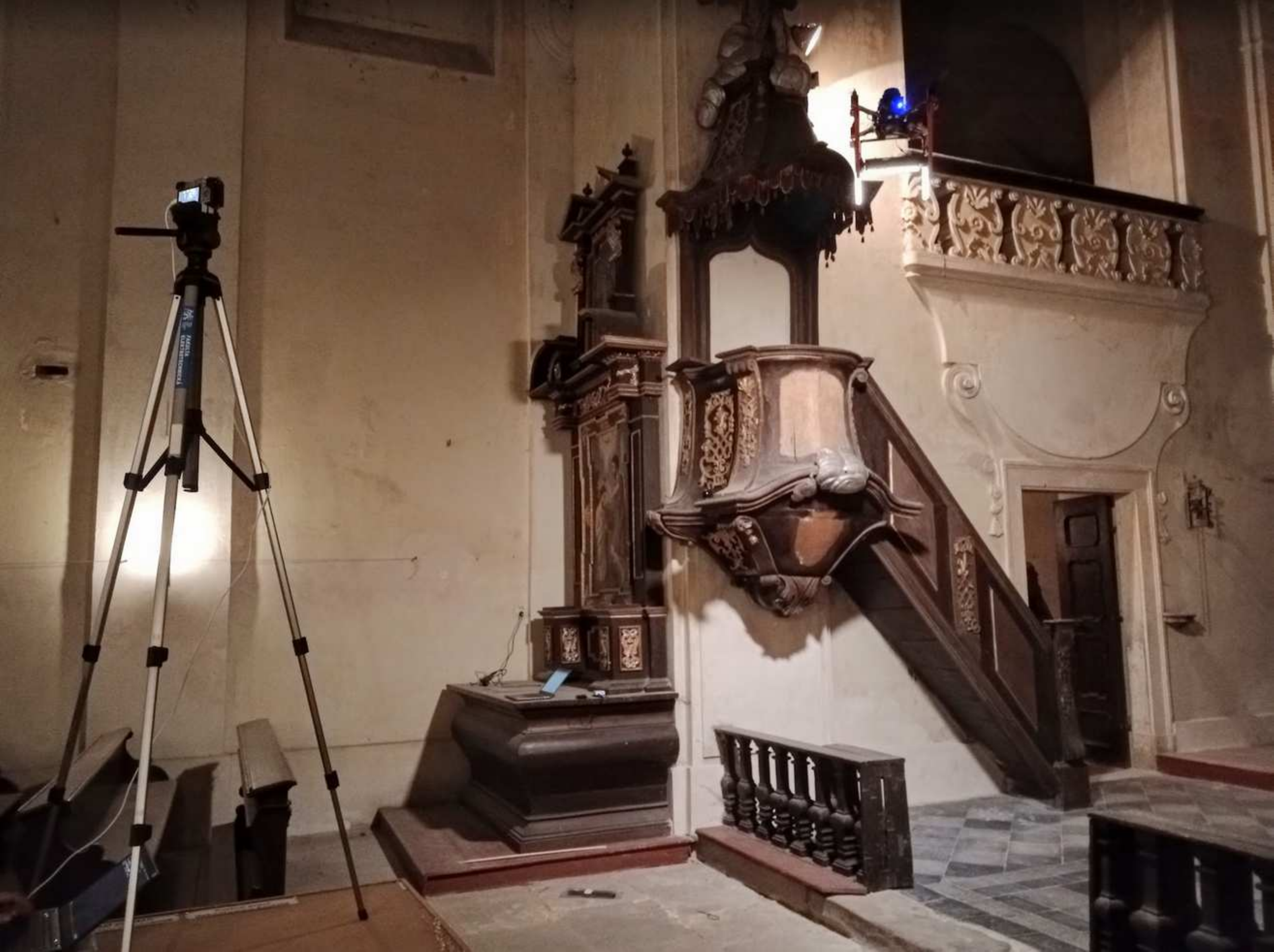}
    \label{fig:experiment_tripod}
  }
  \vspace{-0.1cm}
  \caption[]{Comparison of the normal map obtained with SPPA (12 positions) with lighting vectors computed by H-RTI method (a) and with lighting vectors computed based on the pose of UAV (b) obtained by the application-tailored localization system\hbox{{\cite{icra20petracek}}}. The setup for this experiment is shown in (c).}
  \vspace{-0.5cm}
  \label{fig:real_exp_normals_comparison}
\end{figure}

{For comparison of the proposed approach and H-RTI, results of the RTI method in the form of normal maps are presented in~\hbox{\autoref{fig:real_exp_normals_comparison}}.  These two normal maps were obtained with lighting vectors computed by H-RTI method and with lighting vectors computed based on the pose of UAV obtained by the application-tailored localization system.
  Although the ground truth measurement is not available, we can, based on the known structure of the pulpit, claim that the results obtained with H-RTI method are more precise especially in the surroundings of the reflective ball.

  However, the proposed method has an undeniable advantage in realization of the RTI method in hardly accessible places. Moreover, under the condition of sufficiently precise localization, which is achieved by the applied localization system\hbox{~\cite{icra20petracek}}, the determination of lighting vectors is more precise than its detection from reflections on the ball, which cannot be placed directly in the center of a scanned object.
  The main drawbacks with respect to manually performed RTI lie in the inability to eliminate any camera motion. This issue is partially solved by the image registration process, however, on the high level of details, the imperfections of the alignment can cause unsharpness in images generated from the PTM representation. The camera motion, together with the high exposure time required in dark conditions, also causes the blur in images taken by the camera. However, this problem can be suppressed by the mechanical stabilization of the camera or by use of light source with higher power output, which enables the reduction of exposure time.
}
\vspace{-0.6cm}

%% file: fig/rti_quantiles.tex
\definecolor{amber}{rgb}{1.0, 0.75, 0.0}
\begin{tikzpicture}[font=\small]
  \begin{axis}[ 
    width=0.9\columnwidth,
    height=0.5\columnwidth,
    grid=major,
    legend style={font=\scriptsize},
    legend pos=north west,
    legend cell align={left},
    xlabel= n-th percentile (-),
    ylabel=distance ratio (-),
    xlabel style={
    yshift=1.5ex},
    ytick={0.75, 1.0, 1.25, 1.5, 1.75, 2.00},
    yticklabels={0.75, 1.00, 1.25, 1.50, 1.75, 2.00}]
  \addplot [mark=*,  color=blue, mark options={color=blue, scale=0.7}] table [y=quantile_ratio, x=quantile]{data/quantiles_fib_lkh.txt};
  \addplot [mark=*, color=red, mark options={ fill=red,scale=0.7}] table [y=quantile_ratio, x=quantile]{data/quantiles_fib_mine.txt};
  \addplot [mark=*, color=amber, mark options={ fill=amber, scale=0.7}] table [y=quantile_ratio, x=quantile]{data/rti_quantiles.txt};
  \addlegendentry{LKH vs. Fib-LKH} 
  \addlegendentry{SPPA vs. Fib-LKH} 
 \addlegendentry{SPPA vs. LKH}
  \end{axis}
\end{tikzpicture}

%% file: fig/exp_rti_follower_yaw.tex
{\scalefont{1.5}
  \begin{tikzpicture}[font=\LARGE]
  \begin{axis}[ 
    scale only axis=true,
    width=6cm,
    height=2cm,
    grid=major,
    ylabel= $\varphi_2\,(rad)$,
    xlabel= Time $(s)$,
    xmin=0.0,
    xmax=360.0,
    ]
    \addplot [mark=none, color=blue, ultra thick] table [y=yaw, x=time]{data/rti_experiment_trajectory_follower.txt};
  \end{axis}
\end{tikzpicture}
}

%% file: fig/exp_rti_follower_pitch.tex
{\scalefont{1.5}
  \begin{tikzpicture}[font=\LARGE]
  \begin{axis}[ 
    scale only axis=true,
    width=6cm,
    height=2cm,
    grid=major,
    xlabel= Time $(s)$,
    xmin=0.0,
    xmax=360.0,
    ylabel= $\xi_2\,(rad)$]
    \addplot [mark=none, color=blue, ultra thick] table [y=pitch, x=time]{data/rti_experiment_trajectory_follower_pitch.txt};
  \end{axis}
\end{tikzpicture}
}

%% file: fig/normals_quantitative_comparison.tex
\definecolor{amber}{rgb}{1.0, 0.75, 0.0}
\begin{tikzpicture}[font=\small]
  \begin{axis}[ 
    width=0.95\columnwidth,
    height=0.43\columnwidth,
    grid=major,
    ytick={0.02, 0.04, 0.06, 0.08},
    yticklabels={0.02, 0.04, 0.06, 0.08},
    xlabel= $\sigma$ (-),
    ylabel= $\delta$ (rad),
    xlabel style={
    yshift=1.8ex},
    scaled y ticks = false]
  \addplot [mark=*,  color=blue, mark options={color=blue, scale=0.7}] table [y=avg_rad, x=stddev]{data/normal_maps_comparison.txt};
  \end{axis}
\end{tikzpicture}

%% file: chapters/conclusion.tex
\section{CONCLUSION}
The method for the realization of Reflectance Transformation Imaging with the use of a team of autonomous cooperative UAVs is described in this paper. The method is designed for two multi-rotor UAVs equipped with a camera and light source that are capable of self-localization within a given map of an environment. Three approaches to the generation of sequences of RTI positions are presented, but only one was approved by representatives of the heritage institute for its deployment in historical objects. This solution includes self-designed methods for generation of human-predictable trajectories to enable simple monitoring of correct behavior of particular UAVs by safety pilots, while preserving an effort to generate short trajectories. The compromise between these two criteria enables the safe deployment of the system in real-world scenarios. The main advantage of the proposed solution in comparison to already existing methods is the ability to perform the RTI scanning procedure in places that are hardly accessible or even inaccessible to humans.

The proposed approach was integrated into the system for documentation of historical buildings proposed in~\cite{saska17etfa} and its practical applicability was tested in numerous experiments in interiors of churches in realistic simulator Gazebo and within the real experiment in Church of St. Mary Magdalene in Chlum\'{i}n. Outputs of these tests were evaluated by experts from the field of historical science, who found the results comparable with the results produced by already existing methods, which are limited to accessible locations.

%% file: main.bbl
\begin{thebibliography}{10}
\providecommand{\url}[1]{#1}
\csname url@samestyle\endcsname
\providecommand{\newblock}{\relax}
\providecommand{\bibinfo}[2]{#2}
\providecommand{\BIBentrySTDinterwordspacing}{\spaceskip=0pt\relax}
\providecommand{\BIBentryALTinterwordstretchfactor}{4}
\providecommand{\BIBentryALTinterwordspacing}{\spaceskip=\fontdimen2\font plus
\BIBentryALTinterwordstretchfactor\fontdimen3\font minus
  \fontdimen4\font\relax}
\providecommand{\BIBforeignlanguage}[2]{{%
\expandafter\ifx\csname l@#1\endcsname\relax
\typeout{** WARNING: IEEEtran.bst: No hyphenation pattern has been}%
\typeout{** loaded for the language `#1'. Using the pattern for}%
\typeout{** the default language instead.}%
\else
\language=\csname l@#1\endcsname
\fi
#2}}
\providecommand{\BIBdecl}{\relax}
\BIBdecl

\bibitem{Mytum2018}
H.~Mytum \emph{et~al.}, ``The application of reflectance transformation imaging
  (rti) in historical archaeology,'' \emph{Historical Archaeology}, vol.~52,
  no.~2, pp. 489--503, 2018.

\bibitem{underwater_rti}
D.~Selmo \emph{et~al.}, ``Underwater reflectance transformation imaging: a
  technology for in situ underwater cultural heritage object-level recording,''
  \emph{Journal of Electronic Imaging}, vol.~26, no.~1, pp. 1--18, 2017.

\bibitem{miles_pitts_pagi_earl_2014}
J.~Miles \emph{et~al.}, ``New applications of photogrammetry and reflectance
  transformation imaging to an easter island statue,'' \emph{Antiquity},
  vol.~88, no. 340, pp. 596--–605, 2014.

\bibitem{Mytum2017ReflectanceTI}
H.~Mytum \emph{et~al.}, ``Reflectance transformation imaging (rti) : Capturing
  gravestone detail via multiple digital images,'' in \emph{Association for
  Gravestone Studies Quarterly}, 2017.

\bibitem{rti_conservation}
J.~Valcarcel~Andrés \emph{et~al.}, ``Applications of reflectance
  transformation imaging for documentation and surface analysis in
  conservation,'' \emph{Internetional Journal of Conservation Science}, no.~4,
  pp. 535--548, 2013.

\bibitem{Saunders:2017:RTI:3136628.3136726}
D.~Saunders \emph{et~al.}, ``Reflectance transformation imaging and imagej:
  Comparing imaging methodologies for cultural heritage artefacts,'' in
  \emph{EVA}, 2017.

\bibitem{7467687}
Y.~H. {Kim} \emph{et~al.}, ``Reflectance transformation imaging method for
  large-scale objects,'' in \emph{CGiV}, 2016.

\bibitem{rti_highlight}
A.~Cosentino, ``Macro photography for reflectance transformation imaging: A
  practical guide to the highlights method,'' \emph{e-conservation Journal},
  no.~1, pp. 70--85, 2013.

\bibitem{fibonacci_lattice}
R.~Swinbank \emph{et~al.}, ``Fibonacci grids: A novel approach to global
  modelling,'' \emph{Quarterly Journal of the Royal Meteorological Society},
  vol. 132, pp. 1769 -- 1793, 2006.

\bibitem{saska17etfa}
M.~{Saska} \emph{et~al.}, ``Documentation of dark areas of large historical
  buildings by a formation of unmanned aerial vehicles using model predictive
  control,'' in \emph{IEEE ETFA}, 2017.

\bibitem{roberts:2017}
M.~Roberts \emph{et~al.}, ``Submodular trajectory optimization for aerial 3d
  scanning,'' in \emph{IEEE ICCV}, 2017.

\bibitem{5979939}
J.~P. {Fentanes} \emph{et~al.}, ``Algorithm for efficient 3d reconstruction of
  outdoor environments using mobile robots,'' in \emph{IEEE IROS}, 2011.

\bibitem{6696481}
B.~{Adler} \emph{et~al.}, ``Finding next best views for autonomous uav mapping
  through gpu-accelerated particle simulation,'' in \emph{IEEE/RSJ IROS}, 2013.

\bibitem{4399581}
P.~S. {Blaer} \emph{et~al.}, ``Data acquisition and view planning for 3-d
  modeling tasks,'' in \emph{IEEE/RSJ IROS}, 2007.

\bibitem{7991427}
M.~{Nieuwenhuisen} \emph{et~al.}, ``Chimneyspector: Autonomous mav-based indoor
  chimney inspection employing 3d laser localization and textured surface
  reconstruction,'' in \emph{IEEE ICUAS}, 2017.

\bibitem{hallermann}
N.~Hallermann \emph{et~al.}, ``Vision-based monitoring of heritage monuments
  – unmanned aerial systems (uas) for detailed inspection and high-accurate
  survey of structures,'' in \emph{STREMAH}, 2015.

\bibitem{8598942}
H.~{Qin} \emph{et~al.}, ``Autonomous exploration and mapping system using
  heterogeneous uavs and ugvs in gps-denied environments,'' \emph{IEEE
  Transactions on Vehicular Technology}, vol.~68, no.~2, pp. 1339--1350, 2019.

\bibitem{7152283}
Y.~{Lyu} \emph{et~al.}, ``Simultaneously multi-uav mapping and control with
  visual servoing,'' in \emph{IEEE ICUAS}, 2015.

\bibitem{saska2014jfr}
M.~Saska \emph{et~al.}, ``{Coordination and Navigation of Heterogeneous MAV \&
  UGV Formations Localized by a "hawk-eye"-like Approach Under a Model
  Predictive Control Scheme},'' \emph{International Journal of Robotics
  Research}, vol.~33, no.~10, pp. 1393--1412, 2014.

\bibitem{spurny_mmar16}
V.~Spurn\'{y} \emph{et~al.}, ``Complex manoeuvres of heterogeneous mav-ugv
  formations using a model predictive control,'' in \emph{MMAR}, 2016.

\bibitem{saska_ras15}
M.~Saska \emph{et~al.}, ``Predictive control and stabilization of nonholonomic
  formations with integrated spline-path planning,'' \emph{Robotics and
  Autonomous Systems}, 2015.

\bibitem{baca_jfr18}
T.~B\'{a}\v{c}a \emph{et~al.}, ``Autonomous landing on a moving vehicle with an
  unmanned aerial vehicle,'' \emph{Journal of Field Robotics}, vol.~36, pp.
  874--891, 2019.

\bibitem{ptm_one}
T.~Malzbender \emph{et~al.}, ``{Polynomial Texture Maps},'' in \emph{{CGIT}},
  2001.

\bibitem{baca2018mpc}
T.~B\'{a}\v{c}a \emph{et~al.}, ``Model predictive trajectory tracking and
  collision avoidance for reliable outdoor deployment of unmanned aerial
  vehicles,'' in \emph{IEEE/RSJ IROS}, 2018.

\bibitem{icra20petracek}
P.~Petr\'{a}\v{c}ek \emph{et~al.}, ``Dronument: Reliable deployment of unmanned
  aerial vehicles in dark areas of large historical monuments,'' \emph{IEEE
  RA-L}, In Press: Accepted for publication on January 9, 2020.

\bibitem{Saff1997}
E.~B. Saff \emph{et~al.}, ``Distributing many points on a sphere,'' \emph{The
  Mathematical Intelligencer}, vol.~19, no.~1, pp. 5--11, 1997.

\bibitem{lkh}
S.~Lin \emph{et~al.}, ``An effective heuristic algorithm for the
  traveling-salesman problem,'' \emph{Operations Research}, vol.~21, no.~2, pp.
  498--516, 1973.

\bibitem{DCCS06}
M.~Dellepiane \emph{et~al.}, ``High quality ptm acquisition: Reflection
  transformation imaging for large objects,'' in \emph{International Symposium
  on VAST}, 2006.

\end{thebibliography}
